\documentclass[a4paper,12pt]{report} 
\usepackage[top=1.75in, bottom=1.55in, left=1.5in, right=1.5in]{geometry}
\usepackage{color}
\usepackage{graphicx}
\graphicspath{{./images/}} 
\usepackage{hyperref} 
\usepackage{pifont} 
\usepackage{array} 

\usepackage{float}
\usepackage{subcaption}

\usepackage{amsmath,amsfonts}

\usepackage{nomencl}
\makenomenclature

\usepackage{acronym}

\usepackage{etoolbox}
\renewcommand\nomgroup[1]{%
  \item[\bfseries
  \ifstrequal{#1}{P}{Physics Constants}{%
  \ifstrequal{#1}{N}{Number Sets}{%
  \ifstrequal{#1}{O}{Other Symbols}{
  \ifstrequal{#1}{A}{Abbreviations}{%
  \ifstrequal{#1}{R}{Roman Symbols}{%
  \ifstrequal{#1}{G}{Greed Symbols}{}
  }}}}}%
]}

\usepackage{changepage}
\newlength{\offsetpage}
\newenvironment{widepage}{\begin{adjustwidth}{-\offsetpage}{-\offsetpage}%
    \addtolength{\textwidth}{2\offsetpage}}%
{\end{adjustwidth}}

\usepackage[style=ieee]{biblatex}
\begin{document}




\begin{titlepage}
    
    \begin{center}

		\begin{minipage}{0.2\textwidth}%
		\includegraphics[width=2.4cm]{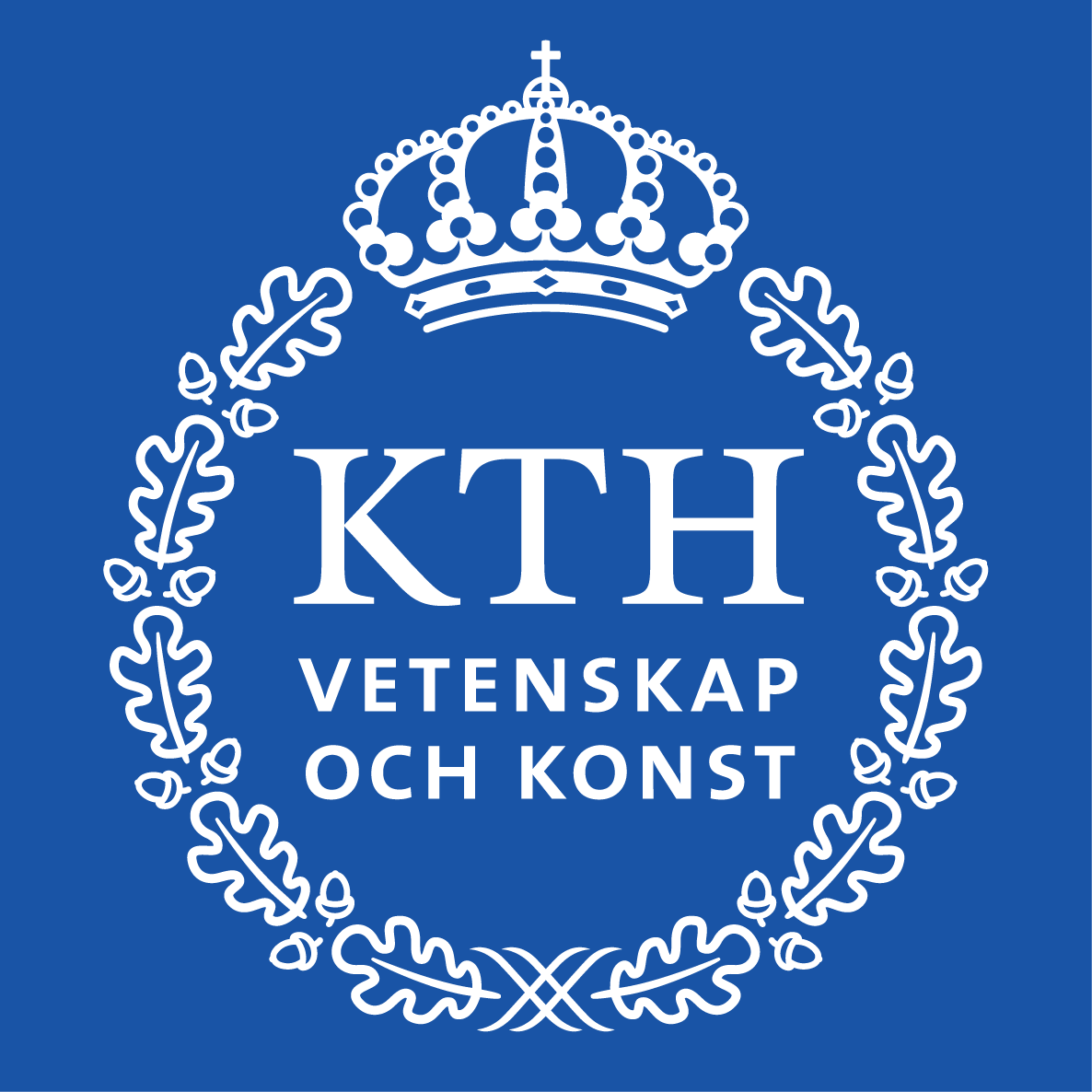}%
		\end{minipage}\hfill
		\begin{minipage}{0.7\textwidth}%
			\LARGE Using Twitter Attribute \\Information to Predict \\Stock Prices
		\end{minipage}%

        \vfill
        \Large
		Roderick Karlemstrand \\Ebba Leckström

		\large
        School of Electrical Engineering and Computer Science

		KTH Royal Institute of Technology

		\vfill

		Supervisor: Bengt Pramborg

		Examiner: Anders Västberg

		\vfill

        \normalsize
		A Thesis Submitted for the Degree of\\
		Bachelor of Engineering in Electronics and Computer Engineering,\\
		Bachelor of Science in Information and Communication Technology

		Stockholm, 2021

    \end{center}
    
\end{titlepage}

\thispagestyle{empty}

\null\vfill
\noindent
Copyright \textcopyright\ 2019 -- 2021 Roderick Karlemstrand \& Ebba Leckström\\
All rights reserved. The figures and images in this thesis may not be reused, reproduced, adapted or redrawn.
\newpage

\pagenumbering{roman}
\setlength\parskip{0pt} 
\setcounter{tocdepth}{1} 

\vspace*{0.2\textheight}
{\itshape "Sometimes when you innovate, you make mistakes. It is best to admit them quickly and get on with improving your other innovations."}\bigbreak
\hfill --- Steve Jobs

\newpage

\section*{Abstract}

Being able to predict stock prices might be the unspoken wish of stock investors. Although stock prices are complicated to predict, there are many theories about what affects their movements, including interest rates, news and social media. With the help of Machine Learning, complex patterns in data can be identified beyond the human intellect. In this thesis, a Machine Learning model for time series forecasting is created and tested to predict stock prices. The model is based on a neural network with several layers of \ac{LSTM} and fully connected layers. It is trained with historical stock values, technical indicators and Twitter attribute information retrieved, extracted and calculated from posts on the social media platform Twitter. These attributes are sentiment score, favourites, followers, retweets and if an account is verified. To collect data from Twitter, Twitter's API is used. Sentiment analysis is conducted with \ac{VADER}. The results show that by adding more Twitter attributes, the \ac{MSE} between the predicted prices and the actual prices improved by 3\%. With technical analysis taken into account, \ac{MSE} decreases from 0.1617 to 0.1437, which is an improvement of around 11\%. The restrictions of this study include that the selected stock has to be publicly listed on the stock market and popular on Twitter and among individual investors. Besides, the stock markets' opening hours differ from Twitter, which constantly available. It may therefore introduce noises in the model. \\

\textit{\textbf{Keywords:}} Stock price prediction, Machine Learning, Deep Learning, Time series prediction, Twitter, Twitter attributes

\newpage

\section*{Sammanfattnning}

Att kunna förutspå aktiekurser kan sägas vara aktiespararnas outtalade önskan. Även om aktievärden är komplicerade att förutspå finns det många teorier om vad som påverkar dess rörelser, bland annat räntor, nyheter och sociala medier. Med hjälp av maskininlärning kan mönster i data identifieras bortom människans intellekt. I detta examensarbete skapas och testas en modell inom maskininlärning i syfte att beräkna framtida aktiepriser. Modellen baseras på ett neuralt nätverk med flera lager av \ac{LSTM} och fullt kopplade lager. Den tränas med historiska aktievärden, tekniska indikatorer och Twitter-attributinformation. De är hämtad, extraherad och beräknad från inlägg på den sociala plattformen Twitter. Dessa attribut är sentiment-värde, antal favorit-markeringar, följare, retweets och om kontot är verifierat. För att samla in data från Twitter används Twitters API och sentimentanalys genomförs genom \ac{VADER}. Resultatet visar att genom att lägga till fler Twitter attribut förbättrade \ac{MSE} mellan de förutspådda värdena och de faktiska värdena med 3\%. Genom att ta teknisk analys i beaktande minskar MSE från 0,1617 till 0,1437, vilket är en förbättring på 11\%. Begränsningar i denna studie innefattar bland annat att den utvalda aktien ska vara publikt listad på börsen och populär på Twitter och bland småspararna. Dessutom skiljer sig aktiemarknadens öppettider från Twitter då den är ständigt tillgänglig. Detta kan då introducera brus i modellen.

\newpage

\section*{Acknowledgement}
We would like to express our special thanks of gratitude to Prof. Anders Västberg for the guidance in this project as well as Dr Bengt Pramborg for providing comments and suggestions to help us write this thesis. Without them, this thesis would not be possible. Secondly, we would like to thank our families. We are extremely grateful for their love, understanding, caring and support. We are also thankful for the help from Eva Boström and Hans Förnestig. Last but not the least, we want to thank Maria Wäppling, Serge de Gosson de Varennes, Hampus Pettersson and Fredrik Schalling from Sopra Steria for their feedback and help.\\

5 April 2021\\
Stockholm, Sweden

\newgeometry{top=1in} 
\tableofcontents
\restoregeometry
\setlength\parskip{6pt} 

\newpage
\section*{Abbreviations}
\begin{acronym}[MPC] 
\acro{AI}{Artificial Intelligence}
\acro{ML}{Machine Learning}
\acro{RNN}{Recurrent Neural Network}
\acro{LSTM}{Long Short-Term Memory}
\acro{JSON}{JavaScript Object Notation}
\acro{API}{Application Programming Interface}
\acro{VADER}{Valence Aware Dictionary and sEntiment Reasoner}
\acro{MSE}{Mean Squared Error}
\end{acronym}




\chapter{Introduction}
\pagenumbering{arabic}

Algorithmic trading is the process of using a computer program to place buy and sell orders. It is often used in the stock or derivative market. A naive such computer program may place stop-limit orders when the stock prices have gone upwards reaching a pre-defined price, and stop-loss orders work similarly as stop-limit ones. It may also be a neural network model, to which the input is market data, and the output is trading orders.

Although it is commonly accepted that stock prices follow a random walk pattern, some researchers have provided evidence that human emotions and psychology do influence the stock market \cite{bollen2011twitter}, \cite{chandra2008decision}. One online platform where people can express their emotions by posting short messages is Twitter.

Twitter is an online social network platform that has gained much popularity in recent years. The correlation between emotions on Twitter and the stock price changes has been studied widely. Previous studies have shown that it was possible to predict stock prices using sentiment analysis on Twitter. Researchers have made many attempts and achieved better performance by using some variant of a Machine Learning (ML) approach \cite{Sul_Trading_on_Twitter_2014}, \cite{pagolu2016sentiment}. Moreover, \cite{Mao_Twitter_volume_spikes_2013} showed that Twitter volumes and follower counts also provide useful information for price prediction.

This thesis covers a study of the design, implementation and evaluation of a Machine Learning model for algorithmic trading with public emotion information collected and extracted from Twitter.

\section{Background}

To know how stocks will be going, investors have been trying their best to predict the future. News is one of the most commonly used information. Even if it cannot be used to predict what would happen tomorrow because news is often about something that has already happened, those who can react faster may buy or sell a stock or option sooner than others, to make profits or reduce losses. On 6 April 2015, a piece of news about a potential deal that Intel acquiring Altera was published on Twitter. Around this time, 3 158 \textit{call options} in the derivative market were suddenly swept. People believed that a computer program that monitored Twitter bought them within a few seconds. Later that day when the stock price went up, the person behind this computer program exercised the options and made \$2.4 million US dollars in 28 minutes \cite{gandel_2015}. Although we cannot eliminate the possibility of being an insider trade, the power and influence of Twitter are becoming undeniable significant.

One way to explain why the price of Altera stock went upwards is that more people were willing to buy and fewer people were willing to sell. The price of a listed stock fluctuates because of the imbalance between the supply and the demand. However, it is not the root cause, rather than how the world would be like in the future economically. It may be a few minutes later, or a few years later. There are many reasons to own a stock, such as to control the company, to receive dividend and to sell it for a higher price sometime later. All these kinds of investors have one thing in common: they believe that the stocks will perform well in the \textit{future}. The reasons to sell a stock are the opposite.

It is nearly impossible to predict the future so that investors count on news \cite{chang_novel_2012}, which gives insights on how the business is going; other people's opinions, out of instincts; or technical analysis, which reflect what other investors think about the stock in the past, right now and the momentum of the price movement.

\section{Problem}
\label{ch-problem}

Predictions get harder and less accurate when the time horizon becomes longer. For example, the weather forecast is most accurate for the following a few hours, less accurate for the next day to come and we do not usually see a forecast for the next month. Alike the weather forecasting problem, it is considered hard to analyse stock market movements and even harder to predict stock prices over long horizons.

To solve this problem, professional investors and researchers in the finance industry have been studying it for decades and have acquired a good understanding of why stock prices change. But there is one variable, social media, that plays a more and more important role in the stock market. President and CEO of Tesla, Ellon Musk, wrote on Twitter that he was considering taking his company private at \$420 per share \cite{rapier_elon_nodate}. He also confirmed that the funding was "secured". Since the market price back then was much lower than \$420 per share, the Tesla stock went up more than 10\% intraday. It was thereafter suspended and U.S. Securities and Exchange Commission started an investigation of manipulating the stock price against Musk. The market reacted to this news and \$23 billion was wiped off share prices for Tesla.

As more and more evidence coming in, social networks are classified as one of the exogenous variables that can impact stock prices. Consequently, new fields of study such as social network analysis and sentiment analysis are born. This thesis promotes a method to solve the future price prediction problem with Neural Networks. Given a dataset of Tweets, how much useful information can be extracted to help to predict future stock prices?

There are already a great number of researches about social networks' impact on stock prices and Twitter, one of the largest social networks has been well-studied. Earlier studies focused mainly on the text itself, but each tweet contains other valuable information, such as how many people liked and retweeted it. We call them Twitter (tweet) attributes. One tweet that has engaged many other users may have more value than the other tweet that no one responded to. By studying how these attributes influence stock prices, it may be possible to acquire a model with better prediction accuracy and precision.

Previous studies like \cite{HongKeelSul_2014} found a strong correlation between cumulative emotional valence, i.e. public moods on Twitter, and stock returns. Moreover, the correlation between the underlying Twitter data and the stock return has been explored in \cite{Mao_Twitter_volume_spikes_2013}, where they acquired statistically significant gain by incorporating Twitter volume spikes into a Bayesian Classifier. Thus, this thesis tries to extend these studies by taking more tweet attributes into account, together with emotions and technical analysis to predict stock prices. The following research questions were studied:

\emph{Will technical analysis and Twitter attribute information help to predict the stock price?}

If the answer to the question above is yes;

\emph{How much better can a machine learning model become with those mentioned above taken into account?}

\section{Purpose}

For stock market analysts, there is a need to develop new tools that in combination with traditional prediction models will tune the predictions by taking into account factors, that do not directly have their origin in the company itself, but the general public’s perception of the market \cite{bollen2011twitter} and opinions of the studied stock in particular. An attempt to extract more information from a dataset of tweets was conducted and the possibility to acquire better stock price prediction with a limited sized data set was studied. The purpose of this thesis is to promote a Machine Learning approach with Neural Networks to try to solve this problem. Often when a Neural Network model is not accurate enough, it can be solved by training it with a larger dataset. Earlier research like \cite{Sul_Trading_on_Twitter_2014} and \cite{Mao_Twitter_volume_spikes_2013} has used only volume and follower count. This thesis extends their findings to study the performance of stock predictions with a Machine Learning model, using five tweet attributes.

\section{Goals}

The goal of this thesis is to build a computer program that can make future stock prices with information from tweet attributes and technical analysis. To achieve that, it is divided into a series of subgoals:

\begin{enumerate}
  \item To collect relevant tweets and stock historical prices.
  \item To carry out technical analysis on historical prices.
  \item To build a machine learning model to predict next-day stock price.
  \item To Validate that the model's correctness.
  \item To compare the Mean Squared Errors (MSE) of next-day stock price predictions between the two models with and without Twitter attributes.
\end{enumerate}

\subsection{Social Benefits, Ethics and Sustainability}

\ac{ML} is a field of study that gives computers the ability to learn without being explicitly programmed by Arthur Samuel \cite{Samuel_1959}. \ac{AI}, which is a superset of ML, refers to any technique that enables computers to mimic human behaviour. Human beings learn manners and ethics from home and school and what we must do and cannot do regulates by law. Since computer programs do not have the equivalent education and there is not yet a complete law system that regulates them, it is critical to think about the impact of implementing ML models on our society, economy and environment.

Stocks are one the most liquid assets hence it is a popular choice to invest in. On 4 June 2020, American Airlines announced that they would re-open their airlines in July. The stock market responded with a rally of 40\% in one day. However, in the derivate market, the price of a call option went up 3000\%. A trading robot may crawl that news and buy all options within a few seconds, just like how a person earned \$2.4 million by spotting the Altera news and buying all available options.

Human tend to focus on the positive news and greediness drives us to jump into the stock market and place buy and sell orders like fish chasing a shiny bait \cite{noauthor_shiny_2021}. What news does not tell us about is that when someone earns 2000\% of profit in the derivative market, their \emph{conterparts} of the trade lost the corresponding amount of money.

Human beings are irrational about investing due to psychology and that makes us misinterpret information, for example, \cite[pp.~6]{chandra2008decision}. Algorithmic trading makes decisions based on the combination of market data and different triggers, to name a few. A stop-loss or a buyback would happen based on carefully crafted mathematical or Machine Learning models to resist human fear and greediness \cite[pp.~25]{chandra2008decision}. The model proposed in this thesis will help to make well-thought-out decisions and to manage the risk of losing money better, which in turn will make a positive and sustainable impact on the private economy and a better society.

When it comes to ethical aspects of AI, people have never stopped discussing it. Sure it can do good things, but if AI made a bad decision that caused economic loss, damage, injuries or casualties, who would be responsible for that? A typical example is Microsoft built a chatbot in 2016 and it interacted with people on Twitter. It took less than 24 hours for the internet to turn the innocent robot into a full racist \cite{vincent_twitter_2016}. For algorithmic stock trading, a computer program in production may be hacked, or has some bug in it, which can result in placing many wrong orders. Some would argue that AI could reduce human error and therefore better than humans. In this way, a great amount of human work will be replaced by AI and many people will lose their jobs \cite{manyika_ai_nodate}. But on the other hand, industries in developed countries have been more and more automated and freed human beings from tough labours.

Training a Machine Learning model often requires a significant amount of computing power hence it consumes a large amount of energy. To help saving energy and increase speed, the training process is usually parallelised using matrix multiplications. It is usually conducted on graphic cards (GPUs). A GPU-accelerated workbench comes at a great economic cost \cite{Garcia-Martin2019}. However, a trained model can be run on a small neural engine like a mobile phone. Considering this distributed use case, the amortised energy consumption down may be as low as running an iPhone app \cite{sagar_does_2019}.

\section{Research Methods and Methodology}

This study is a quantitative research using mathematics and experiments \cite{HakanssonAnne2013PoRM}. Financial data is used to build a machine learning model. Different parameters such as the number of neurons in each ML-layer, which type of layers to be used, and last but not least the activation function of each ML-layer, are used in our experiment to get acceptable results.

The methods used include collecting data from Twitter and Yahoo Finance, using existing mathematical models for technical analysis, as well as building and training an ML model.

\section{Delimitations}

The sampled stocks are a small number of shares listed on the NASDAQ Composite (IXIC). But due to the limitations of the data sets, this study is conducted on the Tesla stock only. The training data is from May 2019 to April 2020. The impact of traditional media and financial reports of the companies are not taken into account. The stock market is volatile and many factors affect the price, such as political, social, economic, technical trends and other stocks’ trends. These factors are either considered since only one exogenous variable is studied.

Because of the limitations of computing power and available time, the amount of data to collect and then process by machine learning algorithms is limited. Therefore the training data might be less than optimal for training the neural network model. Also, fine-tuning of parameters in the model for optimal accuracy might be difficult to perform due to limitations in computing power, time and data.

\section{Diposition}

The structure of the report is as follows. Chapter 1 introduces the background of the topic of this thesis and the problem. Then, the purpose and goals of this thesis are presented and a discussion about social benefits, ethics and sustainability, is conducted. Chapter 2 presents the necessary background that this study is based on. The methods are presented and discussed in Chapter 3. In Chapter 4, the results will be enclosed and explained. Finally, a discussion is conducted and a conclusion is drawn in chapter 5.

\chapter{Background}

This chapter briefly presents the background that this study is based on, such as commonly used financial indicators, sentiment analysis and \ac{ML}. To answer the research question, we started with a literature study. It is presented as related work as follows.

\section{Related Work}

The prices of stocks fluctuate primarily because of the supply and demand of markets, following the theory of microeconomics \cite{wikipedia_supply_2020}. Forecasting stock values are complex due to the non-stationarity, non-linearity and noisy environment which in turn influences the volatilities of stocks \cite{giles_noisy_2001}. A diversity of factors are involved in the predicted value of a stock, such as general economic conditions, political stability, customer value, a company’s customer reviews, traders’ expectations and social media \cite{chang_novel_2012}.

Predicting stock values with mathematical models and computing have been prevalent for the last decades \cite{giles_noisy_2001}. With Machine Learning, huge amounts of data can be processed and analysed to predict patterns of stock trends over time. Patterns and factors which are too complex to identify manually can be done with ML.

Except for mathematical patterns, stock trends can be analysed with theories such as Elliot wave theory, which suggests specific stock movements recur over time \cite{stockcharts_introduction_nodate}. However, theories differ. According to Malkiel \cite{malkiel_random_2003}, stocks can be explained to have a random walk, which means future values cannot be predicted by historical because of their random occurrence. As recently researches shows, stock movement is predictable and it is therefore taken as an assumption in this thesis.

The stock prediction has been carried out since the beginning and by various methods, of pure curiosity of the future or merely greediness. Some with sentiment analysis \cite{bollen2011twitter}, \cite{Lewerentz2018}, \cite{nguyen_shirai_Topic_2015} while some others with ML models \cite{selvin_stock_2017}, and some combine both of them\cite{zhuge_lstm_2017}.

Over the years stock prediction have been calculated by using various methods. The two most prominent techniques are statistical, such as Naive Bayesian Classifier, and neural network-based, such as \ac{RNN}. In 2011, a paper authored by Bollen et al. impressed researchers all over the world by predicting stock prices using sentiment analysis \cite{bollen2011twitter} and this paper is still cited by many today. Thanks to this inspiring research, more studies were made in this problem field using the data extracted from Twitter. \cite{HongKeelSul_2014} found a strong correlation between cumulative emotional valence, i.e. public moods on Twitter, and stock returns. \cite{pagolu2016sentiment} pointed out the drawback in \cite{bollen2011twitter} and suggested two other methods: Word2Vec and N-gram. Moreover, the correlation between the underlying Twitter data and the stock return has been explored in \cite{Mao_Twitter_volume_spikes_2013}, where they acquired statistically significant gain by incorporating Twitter volume spikes into a Bayesian Classifier.

Since the 21st century, computational power has increased dramatically and so does the popularity of ML methods. Artificial Neural Networks, Recurrent Neural Networks, Convolutional Neural Networks and Long Short-Term Memory are some examples. In particular, deep neural networks acquire more accurate results than neural networks \cite{Masaya_Cross_Section_2018}. Adebiyi et al. \cite{adebiyi_comparison_2014} compared a linear statistical method with a non-linear ML method. They concluded that the latter outperformed the former. \cite{selvin_stock_2017} designed a model-free ML approach and \cite{jung-hua_wang_stock_1996} combined statistical method with ML. \cite{liu_stock_2017} could also show that their ML model was superior, as stock market prediction using RNN and sentiment analysis of financial posts lead to more accurate results than using stock data. As the investors believe that historical prices impact future stock prices, Neural Networks with memory, such as LSTM, was evaluated in stock prediction. Lastly, Zhuge et al. \cite{zhuge_lstm_2017} combined LSTM with Twitter emotions to predict stock prices and claimed that they could successfully predict the opening prices of the stocks at their choice.

According to previous studies, stock movements correlate with social media such as news, trending topics online and mood polarity in posts. By involving data from both social media and historical stock data in AI computing, forecasts can be predicted more accurately than when using only historical stock data \cite{nguyen_shirai_Topic_2015}.

The literature study shows that it is beneficial to implement Machine Learning in stock prediction and Twitter has an impact on stock prices as well. This thesis is based on the results and methods of previous studies and attempts to predict stock prices using Twitter attribute information.

\section{Twitter}

\begin{figure}[ht]
  \centering
  \includegraphics[width=0.9 \linewidth]{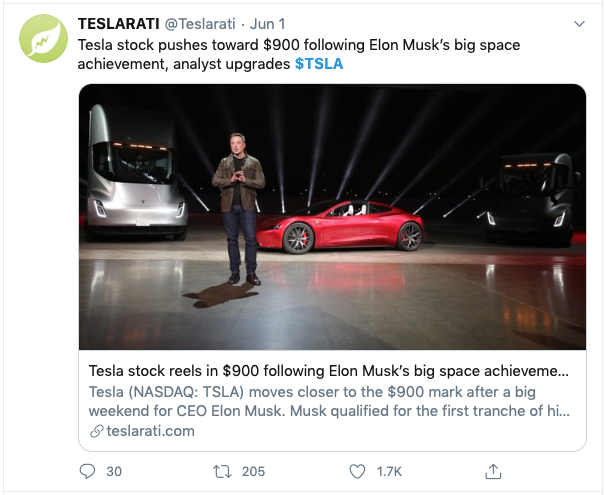}
  \caption{A tweet about Tesla and its stock prices}
  \label{fig:TSLA-tweet}
\end{figure}

Twitter is one of the biggest social media platform offering a micro-blog service, where users can post short messages known as a tweet. Users can react to each other's tweets by liking, re-tweeting, marking them as favourite and commenting on them, to name a few. It is also possible to declare that a tweet is about a certain topic or stock by using hashtags or dollar signs. People can follow others and be followed to keep them updated about what their following users are talking about. Figure \ref{fig:TSLA-tweet} shows one of the results returned when searching for Tesla on Twitter.

From the top there is the Name of the Twitter user/account, the user's Twitter handle and the date of post. It is followed by the text part of the Tweet, limited to 140 characters. Users may also attach a link and the thumbnail of the webpage is displayed if available. The picture, title and preview in the middle of \ref{fig:TSLA-tweet} is the thumbnail of the linked webpage. Lastly, there are four buttons and three numbers at the bottom. The buttons are Reply, Retweet, Favourite and Share. The numbers indicate the number of replies to this tweet, the number of retweets, the number of favourites.

\section{Sentiment Analysis}
\label{ch-sentiment}

Sentiment analysis is known as the process of drawing conceptual meaning and interpret hidden information in a text such as subjective information. In this way, the extracted data can be used in machine learning models or statistical models. When applying sentiment analysis on social media, the text itself has been the main focus. Sul et al. \cite{Sul_Trading_on_Twitter_2014} studied the number of followers of those Twitter users that tweeted about certain stocks. It was found that Twitter users with less than 171 followers which tweeted about a firm, showed a greater impact on the returns of the stock the next trading day, than accounts with more followers.

Sentiment analysis is performed through a systematic approach by algorithms, to extract i.e. polarity, topics, and opinions from the text. Rule-based modelling of language or to compute hidden patterns by artificial intelligence is two methods that can be applied. Sentiment analysis is considered complicated since the syntax and structure of language are not easily summarised and represented with computational models. One of the problems is the ambiguity of words and detecting sarcasm. Words have different meanings depending on the context and use of literary techniques such as irony. Further, text in social media is short and involves emoticons, shortenings, and uppercase letters to emphasize meaning and emotions. \cite{selvin_stock_2017} showed that social media texts can express emotions in a different structure than regular text, which complicates a harmonized sentiment analysis.

\subsection{VADER}

VADER (Valence Aware Dictionary and sEntiment Reasoner) is a lexicon and rule-based sentiment analysis tool written in Python \cite{Vader_github}. VADER consists of a lexicon with a large set of words and emoticons which are labelled according to their semantic orientation with weights. The sum of all words’ weights in a text, is its' resulting polarity. The resulting polarity falls into three different categories: Positive, negative, or neutral. The categories are separated by decimal intervals which makes it possible to evaluate how strong the polarity is per category. VADER can process not only texts but also emojis. It also can detect sarcasm with high probability \cite{Hutto2014}.

\section{Machine Learning}

Machine Learning is a subfield of AI focusing on teaching a machine a specific task, like recognising hand-writings, classifying cats or dogs, re-generating images and time series forecasting, to mimic certain human skills. Besides that, it can also be used to solve problems that are generally hard for human beings, or require great expertise. For example, it may take years of experience for a doctor to be able to identify a bad tumour by looking at the X-ray images, but a computer program can use Machine Learning techniques to reach that level with a much shorter time of training \cite{landhuis_deep_2020}. Even if the computer program cannot replace the doctor, it can help to detect cancer earlier and with better \textit{recall} (fewer false negatives) \cite[pp.~7]{Alpaydin2014}. It can also be used to find patterns in a big amount of data, which are difficult or time-consuming for humans to do the same task.

\begin{figure}[ht]
\noindent\makebox[\textwidth]{%
\includegraphics[width=1\textwidth]{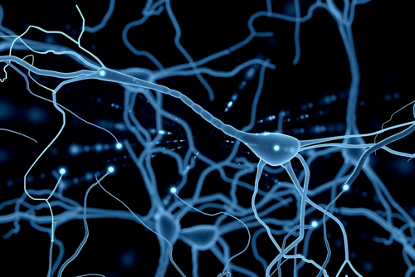}}
\caption{A 3D-rendered Neuron Cell}
  \label{fig:neuron}
\end{figure}

Figure \ref{fig:neuron} illustrates a neuron in the human brain. It is connected to other neurons and can communicate with them using chemical substances. Machine learning and specifically Artificial Neural Networks (ANN) are inspired by how the human brain works fundamentally in decision making and calculation \cite[pp.~13]{Goodfellow2016}. They both are built up of neurons that are transferring signals and deciding which signals to fire off further ahead in the calculation or decision process. Neural networks are built up of different layers with different calculation functions. Some examples of layers are fully connected layers with bias values, dropout layers, and LSTM layers \cite{hochreiter1997long}.

For a neural network to make accurate decisions or perform specific functions with precision, it often requires a large amount of training data from which it will learn to recognise the patterns. There are many training algorithms such as perceptron learning and delta rule learning, Deep Learning and Reinforcement Learning. Initiated by Google, neural networks have been developed and improved dramatically during the last few years and new innovative ideas keep coming and coming \cite{hogarth_state_nodate}.

\subsection{Artificial Neural Network}

The human brain consists of billions of neurons. A neuron cell consists of the cell, dendrites, axon, and synapse \cite{Pannese2015}. When a neuron gives an output, it will travel in form of an electric signal through the axon and reaches the far end. Then, endogenous chemicals called neurotransmitters are released and reach the neuron to which the axon is connected. The amount of neurotransmitter received will trigger electric signals. Now the new neuron will get information and do some processing with this information and output to other neurons \cite{Pannese2015}.

The human brain is an extremely complex organ. The key function is transferring electric signals around the brain which make up different patterns. Long-term potentiation is when the brain, due to its plasticity, creates new paths between synapses. The more often that a path is used, the more prominent it will be thus it will be easier to fall into that connection between synapses. An example of long-term potentiation is when you try to learn something new, new synapses connect through new paths that link them together. Oppositely, old and less-used paths will fade.

For example, a blind person can eventually learn to "see" or at least understand a picture. This by mapping the pixels of real-time images from a camera in form of an array and release electric pulses which strength are based on the greyscale of the pixels on the tongue of the blind patient. After a lot of training the brain has learned which kind of electric pulse matrix indicates which image. And therefore, the patient can see the world by feeling the pulses.

A traditional neural network is a simple multilayer feed-forward artificial neural network. The network usually consists of at least three different layers, input layer, hidden layer, and output layer. Each layer consists of neurons with activation functions. The input goes from the input layer, passes the hidden layers and to the output layer, the network contains no loops between layers.

Depending on how a neural network is designed, the number of hidden layers varies. A few years ago, a few hidden layers are considered deep. But now in 2019, the drawbacks of a deep neural network with only a handful of hidden layers have been discovered and acknowledged. Thus, engineers and researchers keep implementing deeper and deeper neural networks to solve more complex problems. Although some people have implemented deep neural networks with hundreds, thousands of hidden layers, a neural network with a few hidden layers can still outperform non-deep neural networks when it comes to solving relatively simple problems. Thus, there is no single correct value for the depth of architecture, nor is there a consensus about how much depth a model requires to qualify as deep \cite{Goodfellow2016}.

\begin{figure}[ht]
\noindent\makebox[\textwidth]{%
\includegraphics[width=1\textwidth]{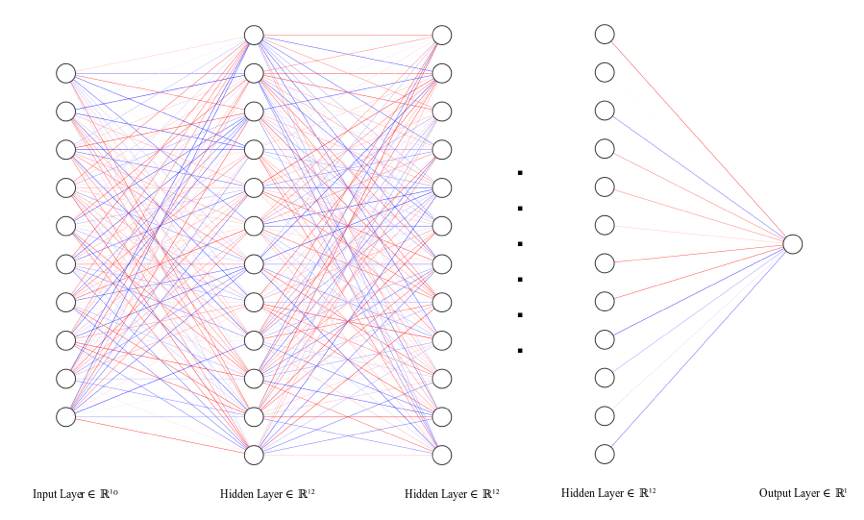}}
\caption{An exampel of a neural network}
  \label{fig:network}
\end{figure}

Figure \ref{fig:network} demonstrates a basic neural network. From the left, there are 10 input neurons which represents the dimension of the input data. They are connected to multiple hidden layers of neurons. The neuron in the output layer will predict a binary classification problem. In this example, negative weights are marked in blue colour while positive weights are shown in red. Some edges are lighter in colour than others. With the help of transparency, the scale of data can be visualized more clearly.

\subsection{Recurrent Neural Network}

Artificial neural networks are inspired by the way the human brain process information. A neural network consists of artificial neurons, of which the topology decides its properties. An RNN differs from a traditional neural network because it introduces feedback loops through the network. Therefore it can be used when the context of the input is important to predict the output. The layers of an RNN are recurrent which means that the current state of a neuron depends on the previous state which gives the neural network a sort of short memory. A recurrent neural network works on sequential data as input, and both the input and output of the network can be sequences of different length which pass through each cell in the network sequentially \cite[pp.~728, 729]{russell2010}.
The system consists of interconnected components called neurons with directed links. Each link to the neuron has weights and biases that are adjusted during the training of the network, to influence the feature strength of the connection and minimize error to be able to adapt to the target output. The activation function decides the output of the layer depending on the weighted sum of inputs, the function can be a threshold or a logistic function \cite[pp.~728, 729]{russell2010}.
RNNs typically use backpropagation to loop information back through the network. It is susceptible to the vanishing gradient problem when the network is too deep. The vanishing gradient problem might appear when the gradient descent algorithm aims to minimize the cost function and slows down the learning speed of the network. This is problematic in deeper networks where the information to process becomes less in every time step of the training process thus the value becomes too small \cite[pp.~733]{russell2010}.

\subsection{Training}
A neural network needs to be fed massive amounts of training data to be qualified for mastering its purpose. From start, a neural network is empty and needs a training algorithm to be properly trained for its purpose. The algorithms are sets of logic that instruct the neural network what tasks and how to perform them. Typical Machine Learning algorithms for Neural Networks are Feedforward and Back Propagation. Each training pass is referred to as an epoch \cite[pp.~733]{russell2010}.

\subsection{Cost Function}

Within the field of Machine Learning, a cost function is a function of the error between computed and true values for an instance of data. A popular method for measuring the error is for example Mean Squared Error (MSE). The value of the cost function after an epoch is referred to as loss \cite[pp.~713]{russell2010}.

\subsection{Feedforward Propagation}

The algorithm has two steps, Feedforward Propagation and Back Propagation. Feedforward propagation first takes the input data into the network and feed it through the first layer. The algorithm feeds the output data from the previous layer through the current layer by multiplying it with the Weight vector of the edge. It then continues feeding the output data into all the hidden layers, one at a time until it reaches the last hidden layer. Lastly, the output layer receives output data from the last hidden layer, processes it, and delivers the new data to the outside of the neural network \cite[pp.~729]{russell2010}.

\subsection{Back Propagation}

As soon as the neural network has computed the final output(s), this value is compared with the true value(s) that is/are found in the training data set. The error is the difference between the predicted value and the true value. The error is what the network needs to "learn". Usually, the network takes a small step towards the optimal solution. And this step size is known as the learning rate. The error is propagated back to the first layer and the weights of the layers get updated \cite[pp.~733]{russell2010}, \cite{ng_cs229:_2019}.

\subsection{Learning Curve}

Two of the most known problems in machine learning are underfitting and overfitting. If the model is too simple, it cannot learn the patterns of the data. Another extreme is called overfitting, which means the model "tries too hard" to learn the training data set and performs therefore badly on the test data set. Both of these cause bad generalisation of the model and it is important to balance them \cite[pp.~705]{russell2010}.

A learning curve is a graph whose x-axis is the size of training size and the y-axis is the error (loss). By sweeping the size of the training set and using the test data set to compute loss we will know if the model has high \textit{bias} or high \textit{variance}. After that, it would be a suitable time to determine what to do next, to balance bias and variance \cite{ng_cs229:_2019}.

\subsection{Long Short-Term Memory}

The prediction methods introduced above have drawbacks because the predictions for short periods can be accurate while predictions far ahead in time might fail. To decide whether to buy or whether to sell a stock, the investors would like to know what the price would be in a month, or often in a year. To solve this problem, Hochreiter and Schmidhuber developed a new machine learning model called Long Short-Term Memory (LSTM) in 1997 \cite{hochreiter1997long}.

\begin{figure}[ht]
  \centering
  \includegraphics[width=0.8 \linewidth]{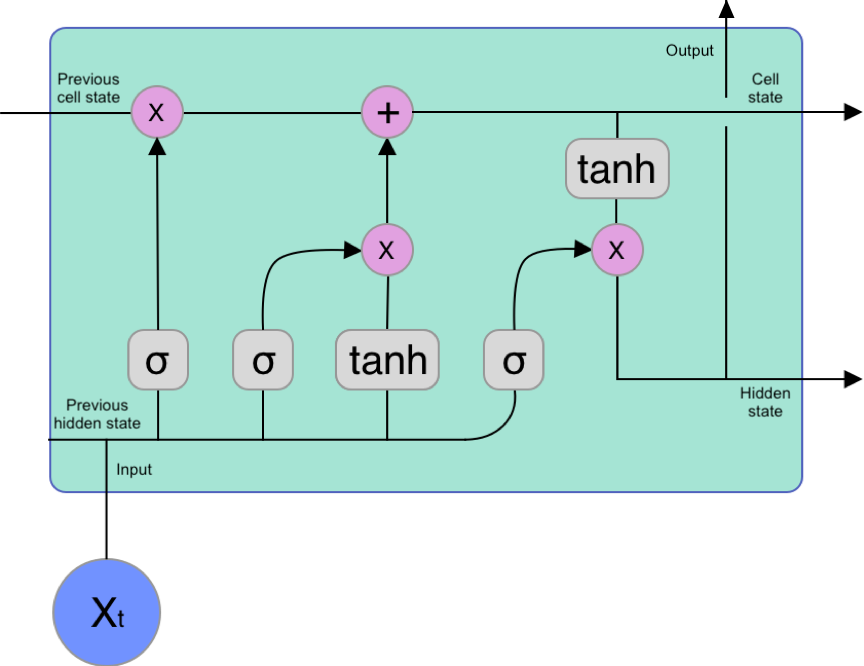}
  \caption{A LSTM cell}
  \label{fig:lstm}
\end{figure}

An LSTM is a type of neuron that has more gates than a traditional recurrent neuron. The fact that LSTM is more complex makes it easier to handle more complex real-world data, like predicting stock prices \cite[pp.~411]{Goodfellow2016}. The underlying architecture of an LSTM cell is illustrated in figure \ref{fig:lstm}.

The inputs are the previous cell state, the previous hidden state and the current input. The symbol $\sigma$ denotes the sigmoid activation function and tanh denotes the hyperbolic tangent function. The + and x denote standard matrix operations. The data flow as follows. The previous hidden state and the input are added together and fed into a sigmoid function. The output goes upwards and is then multiplied with the previous cell state. It is thereafter added with the product of the results from the sigmoid function and the tanh function, shown in the middle of the figure. The final result becomes the current cell state. To produce the output, this cell state is then fed into a tanh function and multiplies with the result from the sigmoid function to the right. The output is also the hidden state to the next LSTM cell.

LSTM tries to mimic the human brain by adding a \textit{forget} function into the cell. The first sigmoid function is referred to as the \textit{forget gate}, as the weights and biases control how much information from history (or the current input) it will forget. In the middle, we have a sigmoid and a tanh function. They are referred to as the \textit{input gate}, and the sigmoid and the tanh function to the right are referred to as the \textit{output gate}. Exact mathematical representation may be found in \cite{hochreiter1997long}. Because of its recurrent property, it can process sequential information over long time dependencies \cite[pp.~410]{Goodfellow2016}.

\section{Technical Analysis}

Technical analysis is a method of studying the stock movement and trading data to predict future prices, in contrast to studying its financial data or news \cite[pp.~1]{Murphy1999}.

\begin{figure}[ht]
  \centering
  \includegraphics[width=0.9 \linewidth]{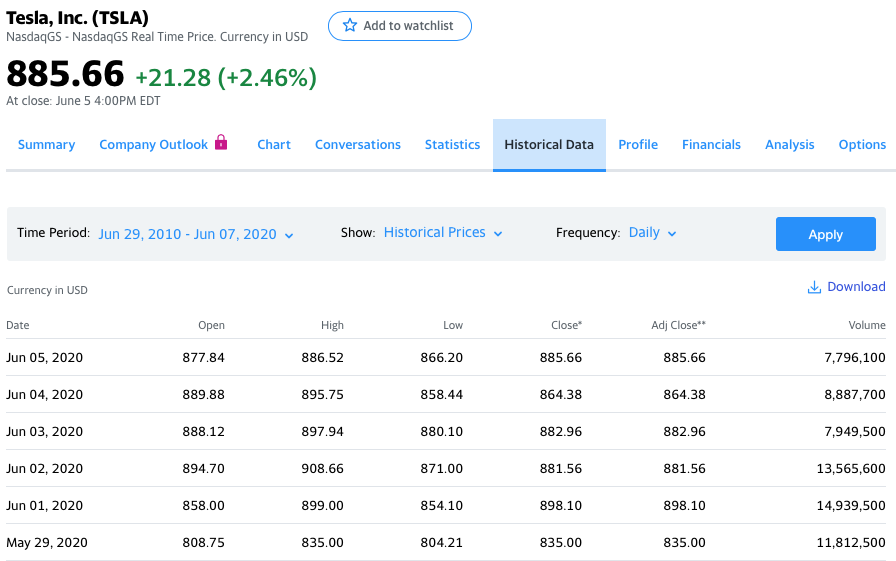}
  \caption{Stock historical data provided by Yahoo Finance}
  \label{fig:yahoo-finance}
\end{figure}

Take Jun 1st 2020 for example in the figure \ref{fig:yahoo-finance}, the stock TSLA opened at 858 USD/share and the price once dropped 0.45\% to 854.1 USD/share, which is the lowest price intraday. Then the stock was traded upwards 7.6\% higher than the day before, up to 899 USD/share and closed that trading day at 898.1 USD/share.

We can see that the buyers were willing to pay more and more price per share. The sellers were not willing to sell their shares at the same price, rather demanding more and more price. This trading pattern may be seen as the market is more united at a \textit{long} position (which means go up). The traded volume, 14 939 500 shares that day may also indicate that the public mood is more optimistic than cautious.

\begin{figure}[ht]
\noindent\makebox[\textwidth]{%
\includegraphics[width=0.7\textwidth]{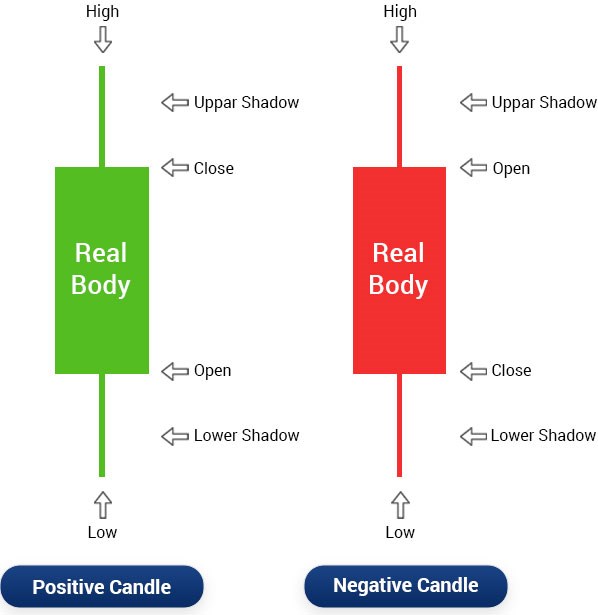}}
\caption{Candle Stick Chart}
  \label{fig:candles}
\end{figure}

Typically, online brokers provide trading software for investors with some technical indicators built-in to perform technical analysis. One of the foundations to this is Candle Stick Chart, shown in figure \ref{fig:candles}.

The chart incorporates the information of stock prices and looks like a candle, hence the name. To the left, we have a \textit{long/positive} candle. The filled part in green is called a body. The stick above and below is called upper shadow and lower shadow. The lowest point and the highest point indicate the lowest price and the highest price in a certain period. The top and the bottom of the body indicates the close and open price in a certain period \cite[pp.~37]{Murphy1999}.

To the right, there is a \textit{short/negative} candle. It works similarly as a positive candle, except for one difference: the top and the bottom of the body indicates the open and close price in a certain period.

\begin{figure}[ht]
\noindent\makebox[\textwidth]{%
\includegraphics[width=1.4\textwidth]{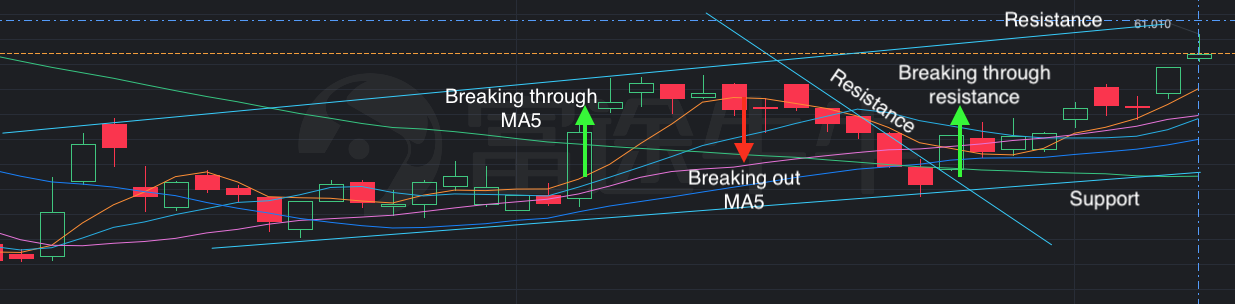}}
\caption{Stock prices and technical indicators}
  \label{fig:trend-channel}
\end{figure}

Now we move on to a 1-day tick chart. Figure \ref{fig:trend-channel} shows a stock price movement within 39 trading days using 39 candle charts, each representing 1 trading day. There are also lines of different colours, such as green, blue, purple, orange, cyan represents SMAs (simple moving averages, see chapter \ref{ch-technical-indicators})

There are also weekly tick charts, monthly tick charts etc. for each security, given that the security has been listed long enough. For short-term investment, trends are usually analysed using the 1-day tick chart.

There are many theories about drawing support lines and resistance lines, as well as how to use them to predict trends. One way of doing that is shown in the figure \ref{fig:trend-channel}. It demonstrates a simple technical analysis. The local maxima and minima are identified and they form a parallelogram which is known as a \textit{trend channel} \cite[pp.~40]{Murphy1999}. It means that the stock will with a high probability to move inside the channel. If the stock \emph{close} price is above the moving average ($n$) line then it is believed that the stock is going up in periods of $n$ and vice versa. Breaking through the resistance level indicates that the stock is about to move upwards, and breaking out the support level indicates the opposite. In this particular case, we can observe that technical analysis successfully predicted stock movement intraday or a few days ahead.


\subsection{Technical Indicators} \label{ch-technical-indicators}

Technical indicators, or technical oscillators, are mathematically calculated signals that are used in technical analysis to predict future stock price movements. Some widely used technical indicators are presented below.

\subsection{Moving Averages}

Simple Moving Average (SMA) is calculated by taking the average of a series of values. Usually, the period ${n}$ is one day and $A_{n}$ is the closing price of the day ${n}$. 5-day-average is written as SMA5 or SMA(5) \cite[pp.~199]{Murphy1999}.

\[ SMA = \frac{A_{1} + A_{2} + ... + A_{n}} {n} \]

\subsection{Bollinger Bands}

Bollinger Bands are a technical analysis tool for generating oversold or overbought signals. The middle band is also used to indicate a trend change. The bands are computed as follows:

\[MB = SMA(n)\]
\[UB = SMA(TP, n) + m * \sigma[TP, n]\]
\[LB = SMA(TP, n) - m * \sigma[TP, n]\]

where:

SMA is simple moving average and n is usually 20 or 30,

MB = Middle band,

UB = Upper band,

LB = Lower band,

TP (typical price) = (Highest price + Lowest price + Close price) / 3,

m is the number of standard deviations (usually 2),

$\sigma[TP, n]$ is the standard deviation over last n periods of TP \cite[pp.~209]{Murphy1999}.





\chapter{Methods}

\begin{figure}[ht]
  \centering
  \includegraphics[width=0.9 \linewidth]{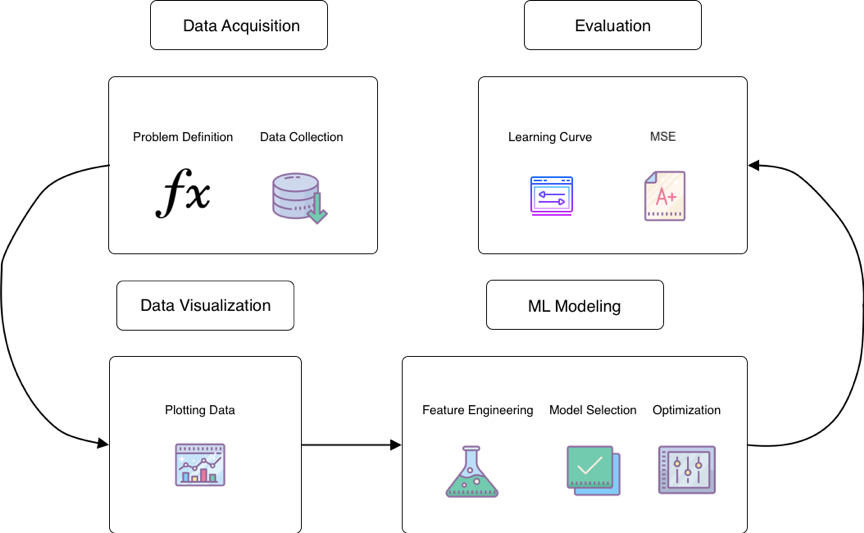}
  \caption{The workflow of this project}
  \label{fig:workflow}
\end{figure}

The chapter presents how the study was carried out. The selected methods are motivated and discussed first, followed by a short explanation of them. References are also provided for those who want to read further. The workflow of this project is shown in \ref{fig:workflow} and will be explained in detail in this chapter.



\section{Choice of methods}

In the literature study, two data collection methods were investigated. The first alternative was to use a Twitter archive that was available for free on the internet. However, it did not contain the complete set of tweets, nor any relevant tweets about stocks. The second alternative was to purchase a Twitter Developer subscription. Since the price was quite high and this project was not sponsored by any company or organisation, the third option was therefore used. Twitter has a free tier of Twitter developer which offers access to historical tweets with limits. To comply with its user license agreement, we decided to customise a Java program that collects tweets with Twitter API.

In the data pre-processing part, Sul et al. \cite{HongKeelSul_2014} utilised one single attribute, Twitter follower count to predict the emph{abnormal return} and promising results were found. Other attributes associated with tweets that were dropped in their pre-processing step are therefore taken into account in this study to answer the research question mentioned in chapter \ref{ch-problem}.

It is also known that financial instruments are widely used to predict stock movements. As presented, previous studies have also implemented such and reached a not great result. These instruments can be easily implemented as an extra \textit{column} in the ML datasets. We believe that investors are affected by psychology \cite{chandra2008decision} hence we would like also to research if the addition of financial instruments helps.

For evaluation, the objective is to find the best hypothesis function (the neural network model) to predict future data. It is therefore often referred to as the objective function or loss function. For regression problems, typical loss functions are mean squared error, mean absolute error and 0/1 loss \cite[pp.~711]{russell2010}. The mean squared error function has a smoother curve. We believe that it will help to reach convergence better because the learning step involving computing the gradient of the loss function. As the gradient is constant for the mean absolute error, the loss value would oscillate around the minimum point.

Other options include Mean Squared Percentage Error which is common in trading because it is generally accepted that the percentage of change matter much more than value changes \cite{bollen2011twitter}, \cite{pagolu2016sentiment}. However, a comparison between multiple data sets is not needed and the purpose of this study is not to make money. Thus, the MSE loss function was chosen.

\section{Development Environment}

This thesis involves software development in Java and Machine Learning in Python. For tweet collection, we used the library Twitter4J which provides the search function and Twitter API authentication. We developed API limit control and parsing the tweet on top of the library.

The data is preprocessed and cleaned in a Python program that was written from scratch with the machine learning library Scikit-learn. The sentiment scores are computed using the software VADER.

As for neural networks, we build our neural networks using the machine learning library Keras. Development and experiments were carried out using PyCharm, a Python IDE (Integrated Development Environment). The Keras library provides fundamental types of "neurons" so that machine learning engineers can focus on designing the architecture of the neural network, tuning hyperparameters and understanding their data. The processing of data as well as the technical analysis was written from scratch using NumPy, which is a library providing numerical functions, like matrix operations. Finally, data was visualised using the library Matplotlib.

\section{Data Collection by Twitter API}


\ac{API} is easily explained as a tunnel to make two computer programs talk to each other.

It is well known in data science that the quality of data is one of the most important factors. We started with exploring data sets that suit our needs and are available publicly, one of them being a website claiming that they had the Twitter archive and they allow a full archive dump (rather than just searching on the website). However, the archive covers only a small fraction of the tweets and we could find only a few tweets per months that were related to our topic.

Therefore, we decided to collect tweets using Twitter's API, which is provided to developers who want to use Twitter data in their projects. Twitter Inc. is a commercial company that offers free microblogging service. In return, Twitter may show ads and sell users data to cover the costs and make a profit. Twitter has gained a huge market and many users use Twitter to interact with each other, hence it becomes more and more popular for researchers to acquire public opinions from it. To gain access to all historical as well as real-time data, a premium subscription is required. Twitter has also restricted the usage with several tiers \cite{twitter_pricing}. Since this project is not sponsored by any company or organisation, the free subscription has been used; It means that only the latest 7-day’s data is available and the total GET requests that can be accepted by Twitter’s API server is limited to 180 requests every 15 minutes.

We developed a Java program to utilise Twitter API to collect the data. We chose to study the stocks AMD (Advanced Micro Devices), Intel, Tesla and Microsoft. The main reason is that these stocks are popular on Nasdaq, and they are volatile. Furthermore, they are well-known so that people are discussing on Twitter all the time. In the program, we use the functions provided by Twitter API to search for tweets under certain constraints including period, as well as observing search rate limit. As mentioned before, the rate limit is 180 requests every 15 minutes. If the limit is reached, we would get an error from Twitter API. Thus, we implemented a rate monitor which pauses the searching progress when the limit is nearly reached. A watchdog is also implemented so that the program will restart if any error is encountered.

\begin{table}[ht]
\caption{Fields in the returned Tweet object used in this study}
\begin{widepage}
\centering
\begin{tabular}{l l l l}
\hline\hline
Field & Description\ & Use case\ \\ [0.5ex] 
\hline
text             & The actual text of the Tweet.                                                                            & \begin{tabular}[c]{@{}l@{}}Keyword extraction and senti-\\ment analysis/classification.\end{tabular} \\
favorite\_count  & \begin{tabular}[c]{@{}l@{}}How many times the tweet \\has been favorited by other \\users.\end{tabular} & \begin{tabular}[c]{@{}l@{}}Can be used to understand \\how popular a Tweet is.\end{tabular}          \\
follower\_count       & \begin{tabular}[c]{@{}l@{}}The number of followers the \\User who posted this Tweet.\end{tabular}   & \begin{tabular}[c]{@{}l@{}}To understand how influential \\this user is.\end{tabular}     \\
retweet\_count & \begin{tabular}[c]{@{}l@{}}The number of retweets of \\the original Tweet.\end{tabular}        & \begin{tabular}[c]{@{}l@{}}To understand how influential \\this Tweet is.\end{tabular}     \\
tweet\_count   & \begin{tabular}[c]{@{}l@{}}Total number of Tweets within \\a certain period.\end{tabular}      & \begin{tabular}[c]{@{}l@{}}To understand how popular a \\certain topic is.\end{tabular} \\

verified\_account & \begin{tabular}[c]{@{}l@{}}If the account is verified by \\Twitter.\end{tabular} & \begin{tabular}[c]{@{}l@{}}To verify the account is the \\"real" account.\end{tabular} \\
\hline
\end{tabular}
\label{table:tweet-obj}
\end{widepage}
\end{table}

After the program sends a search request to Twitter through API, it will catch the result returned from Twitter in a \textit{dictionary} format. A dictionary is a data structure that maps keys to values. Table \ref{table:tweet-obj} shows examples of the fields (parameters associated with a tweet) in the API response. So that the data can be processed later in a machine learning library, we parse the Twitter object and stores the data in \ac{JSON} format.

Inside the API response, the object tweet contains 15 to 30 tweets within the period that the API query requested, and the object query provides the information about whether all of the tweets are received or not. This is essential because the program will request more tweets if the query is \textit{not null} but will stop requesting and stores all data into a JSON file if it is \textit{null}.

As mentioned in previous chapters, different Tweet may have a different impact. For example, a verified account may have more authority to comment on something, and their opinions may be more trusted by other people; Or similarly, if an account is followed by many people, it may reach the wider public and therefore impact the stock prices more. To limit the scope of the project, We chose to extract 5 attributes associated with each Tweet: the sentiment score, number of favourites, number of followers, number of retweets and if an account is verified. The total number of tweets is also taken into account. But since this value is not associated with the Tweet, it is considered an extra feature in the training data set rather than another attribute.


As for stock prices, Yahoo Finance provides free access to full historical data, such as open, high, low and close prices as well as traded volume (no intraday data). As shown in \ref{fig:yahoo-finance}, there are 6 values for each day. The variation of the price intraday may reflect the difference between how desire investors are willing to buy and sell a stock. How this information is used is explained in the next section.

\section{Data Processing}

\subsection{Creating Training Examples}

\begin{figure}[H]
  \centering
  \includegraphics[width=0.9 \linewidth]{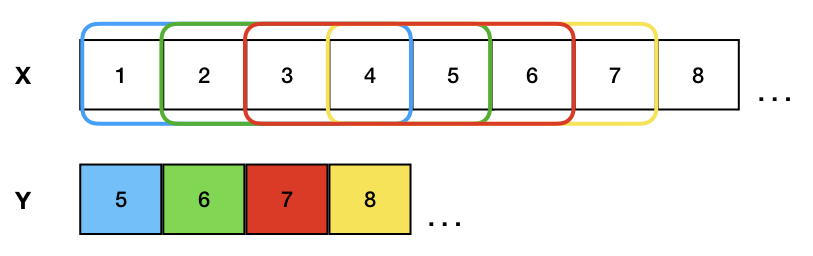}
  \caption{Creating training examples with filter length 5 and step size 1.}
  \label{fig:data-processing}
\end{figure}

Before the collected data can be used to train the neural network model, they usually need to be processed first. The reasons for doing so are many, such as to remove noise or to make it easier for the neural network to be able to extract patterns. For time series prediction problems, a filter is applied to the data to create the training examples. Figure \ref{fig:data-processing} demonstrates this process with a filter length 5. Each colour corresponds to each unique input data to the network. The first 5 elements are taken from the data set. The first 4 elements will be the input (x) and the last element will be the output (y). Next, the filter moves forward with step size 1 and another training example is created. The filter will continue moving forward and this process is repeated until the filter reaches the end of the data set.

During training, the network will learn to use the input data to predict the output data, which means, given a series of 4 consecutive values, the model will output what value it believes the fifth value would be.

\subsection{Adding Technical Indicators}

The moving averages are calculated using the formel described in chapter \ref{ch-technical-indicators} and are added to the data set. When calculating SMA5, 5 values will be used to calculate the average value. The first row of the stock data is 23 May 2019 and to calculate SMA5 for the same day, 4 previous days stock value are needed. Thus, we have taken 20 extra days of stock data to calculate Bollinger bands.

\subsection{Sentiment Scores}

To quantify the sentiment of tweets, a tool called VADER (Valence Aware Dictionary and sEntiment Reasoner) is used. VADER is available as a Python package, \textit{vaderSentiment}. The Python class \textit{SentimentIntensityAnalyzer} inside this package takes text as input and output a \textit{score} object. We created an object of SentimentIntensityAnalyzer and called it with daily tweets. Then we extracted the \textit{weighted compound} score and saves it in a CSV file. The score is finally added to the data set before training.

\section{Neural Network Model}

Earlier studies pointed out that the buyers and sellers use the historical prices as well as the current trend to determine the price they are willing to pay and accept \cite{Jegadeesh2001}, \cite{Edwards2018}. Thus, the historical prices, as well as moving averages, will affect future stock prices. It is therefore considered as a time series forecasting problem, which makes it suitable for applying machine learning.

To reach the goal of the project, a computational model that can compute and combine both statistical patterns, time-oriented patterns and random occurrence in data is necessary. The model should not overfit or under-fit, and must be possible to combine two types of input training data; historical stock- and twitter data, combined with additional features from Twitter data.

Firstly, we collect historical stock price data from Yahoo Finance for up to 10 years. Next, we use several mathematical models that are described in Chapter \ref{ch-technical-indicators}, technical indicators. After this data processing step, the training data for ML consist of 200 rows, which represent 100 trading days. Each row contains market data, such as open price, close price, volume, percentage, as well as sentiment scores of tweets the day before the actual trade day. The sentiment score is calculated using VADER. All these features are added as associations to stock price movement for a single day, which provides now a complete data set. The data set is then split into a training set, a verification set, as well as a test set, which are used for machine learning \cite{ng_cs229:_2019}.

\begin{figure}[H]
\begin{widepage}
  \centering
  \includegraphics[width=0.7\linewidth]{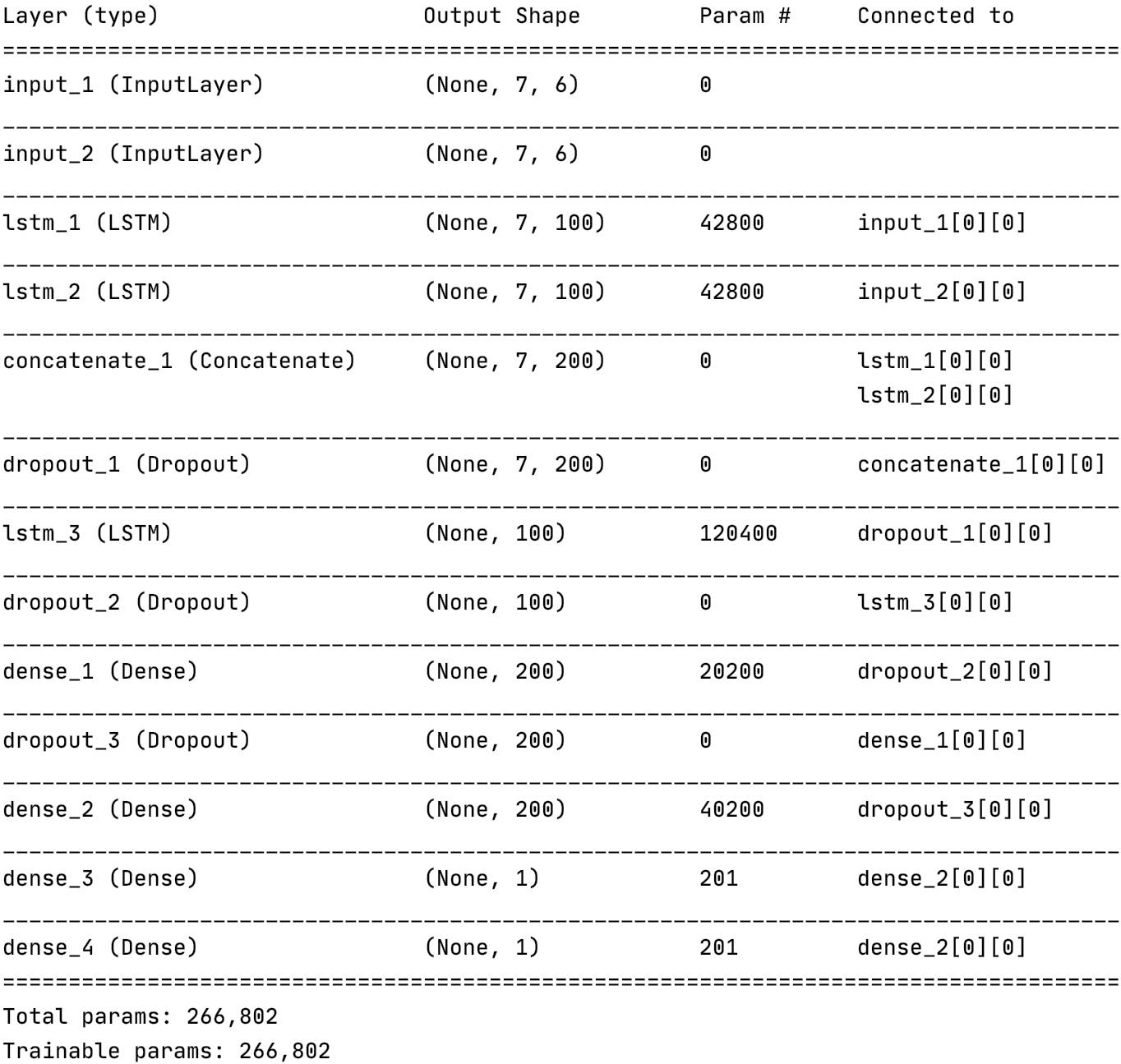}
  \caption{Neural Network Model}
  \label{fig:model}
\end{widepage}
\end{figure}

Figure \ref{fig:model} shows the architecture of our machine learning model. First, the input layer takes series of normalised stock prices, technical indicators etc. as input. Then, the data is fed into the LSTM layers. The LSTM nodes try to extract 100 features from them and output to the next layer \textit{concatenate}. This layer combines all features from input 1 and input 2 and sends them to another LSTM layer. The output is sent to 2 fully connected layers, i.e. the dense layers. Finally, the last dense layer will predict the stock price. To prevent overfitting, dropout layers with a probability of 20\% are added between the connected layers. In the neural network, we use 10 neurons in the input layer, which represent the number of columns in the data set. The hidden layers are specified in the figure. Then output layer predicts the future price and the MSE is computed. For tuning the hyperparameters, we experimented with different step-sizes, different numbers of hidden layers, neuron types, dropout rates and activation functions to acquire an acceptable result.

\newpage

\section{Evaluation of Model}

The performance of the machine learning model is evaluated by comparing the predicted price and the real close price and computing the mean squared error.\\

MSE = $\displaystyle\frac{1}{n}\sum_{t=1}^{n}e_t^2$

where $e$ denotes the error between the ground truth and the predicted value.

The MSE is the average of accumulated error across the whole validation and test data set within every epoch to measure the actual performance of the hypothesis function (i.e. neural network model).

\chapter{Results}

In this chapter, the results of the development of a Java program as well as the machine learning model are presented.

\section{Data Collection}

Since Twitter does not limit the \textit{total} number of requests made in each month (they do limit for premium subscriptions), we ran the data collection program 24/7 to get as much data as possible. One search request usually returns 15 to 30 tweets and we have collected.

The machine learning model takes the close prices each day the US stock market is open as input and iterates through them. To create enough training data, we can of course collect more data, or try to manipulate the existing one. The latter option is normally used in the tech industry, like mirroring, rotating the images in the training dataset. In our case, we have derived more data from it by computing the Bollinger Bands and MAs. More training data was also created by using a sliding window. The window takes a portion of the whole training data and masks the rest. Then, it slides along the training set with a \textit{step size} and repeats the previous step. The step size is a standard training parameter used in ML. We have by trial and error method optimised the \textit{step size} in the training process, which is 7-10 trading days.

To match the same number of trading days, we had to collect the same number of days for tweets. Otherwise, it would throw a list size mismatch error. However, the problem was that trading days are more than 28\% less than tweet days, as well as that if the training days are too few, we would not get any optimal results. That's why we collected tweet over a year and used all of them in our ML model.

The free version of the Twitter developer program offers 180 requests every 15 minutes. A single twitter day usually returns thousands of results, why they were broken down by Twitter and sent to us in batches. After the whole result was received, they would then be saved to the hard drive from computer memory. Sometimes twitter API server was unstable and the collection was terminated in the middle. Sometimes due to the imperfection of our Java program, the program might crash and all unsaved data in memory was lost. Thus, the data collection process had been carried out 24/7 on our local machine. But around 3\% of the data was not collected properly. For the tweet days that were collected successfully, 20,000 - 50,000 tweets per day for Tesla, and around 10,000 - 30,000 per day for Intel, almost 7 GB in total, have been acquired. The tweet attributes, like "Favorites" and "Followers", are also saved. The emojis used on Twitter is encoded in Unicode and can be then processed by VADER. More about emojis is explained in chapter \ref{ch-sentiment}.

From the beginning of this project, we tried to collect tweet about Apple Inc as well as Advanced Micro Devices, Inc. The former one had a great amount of noise in the result since apple can also refer to the fruit. As for the latter one, AMD did not have near the popularity as other tech companies, like Tesla, Nvidia and FAANG (Facebook, Apple, Amazon, Netflix and Google). The reason may also be that most of the people talking about AMD tend to talk more about AMD's product, the Ryzen processor series, than the company itself. Therefore in the machine learning part, only Tesla and Intel tweet are used as the training dataset.

\section{Feature Engineering}

We have chosen to use SMA5 and Bollinger Bands to add features from the perspective of technical analysis to the training data. The Bollinger middle band uses 20 days moving averages of the close prices of the TSLA stock. In total, 4 extra features were extracted and added to the training data set.

\begin{figure}[H]
  \begin{widepage}
    \centering
    \begin{subfigure}[b]{0.4\textwidth}
        \includegraphics[width=\textwidth]{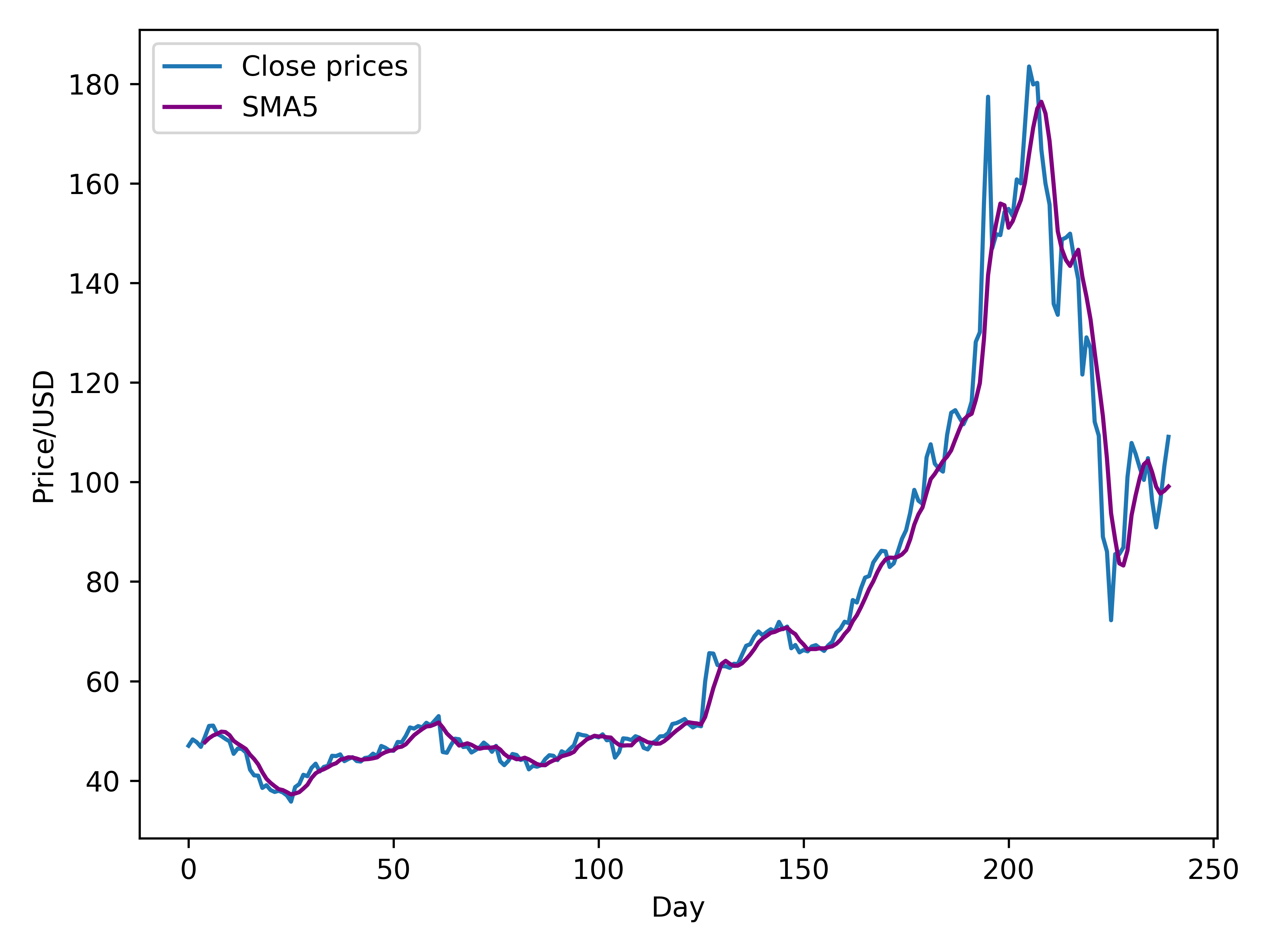}
        \caption{The TSLA stock and SMA5}
        \label{fig:exp2-1-a}
    \end{subfigure}
    ~ 
    \begin{subfigure}[b]{0.4\textwidth}
        \includegraphics[width=\textwidth]{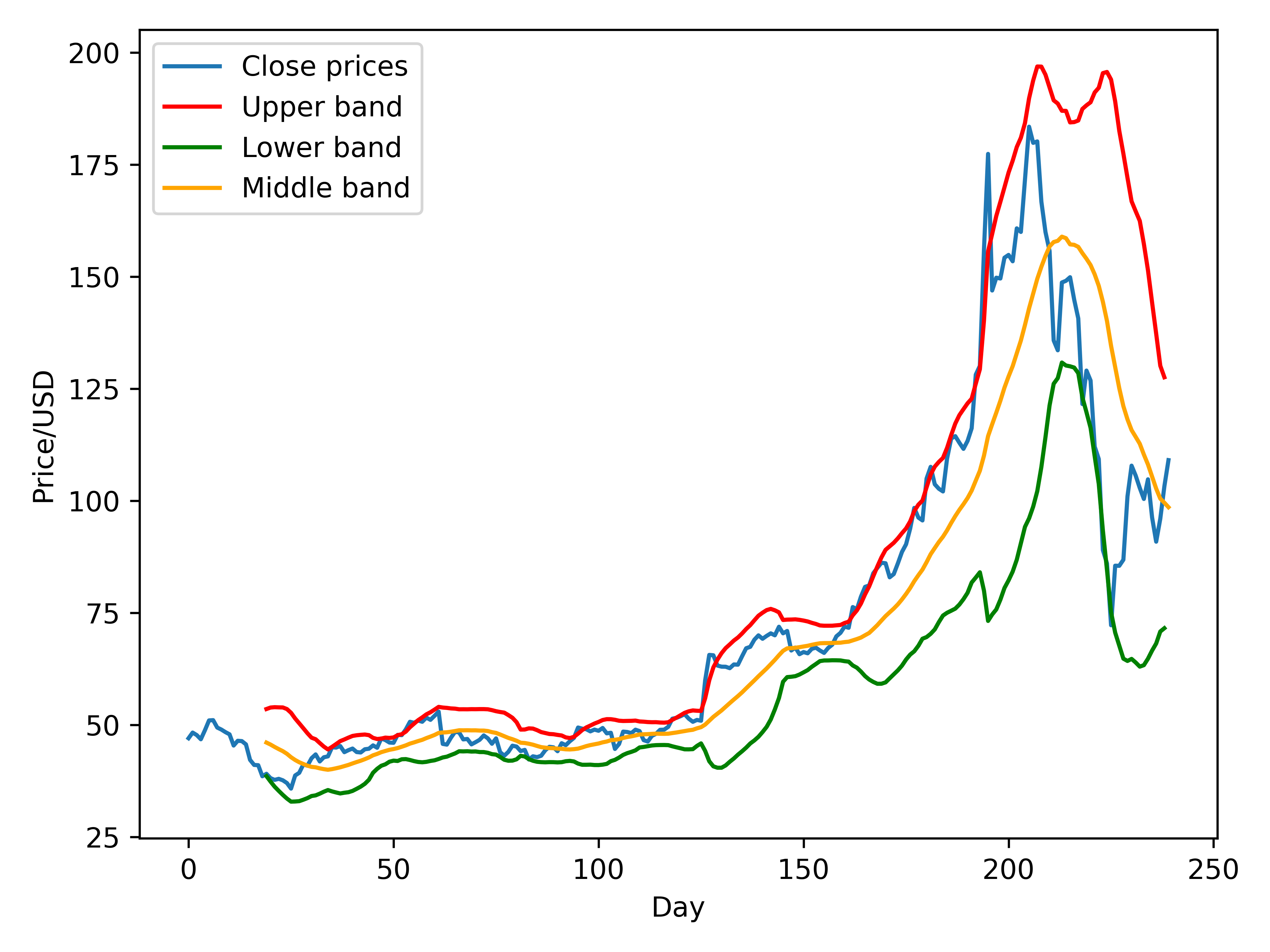}
        \caption{The TSLA stock and Bollinger Band}
        \label{fig:exp2-1-b}
    \end{subfigure}
  \end{widepage}
  \caption{The TSLA stock close prices in USD and technical instruments from 23 May 2019 to 6 April 2020}
  \label{fig:exp2-1}
\end{figure}

\ref{fig:exp2-1-a} shows the stock prices and SMA5, where SMA4 reduces the noise from the stock but keeps the moving momentum in place. \ref{fig:exp2-1-b} shows the Bollinger Bands, which is widely used in technical analysis.

\begin{figure}[H]
  \begin{widepage}
    \centering
    \begin{subfigure}[b]{0.4\textwidth}
        \includegraphics[width=\textwidth]{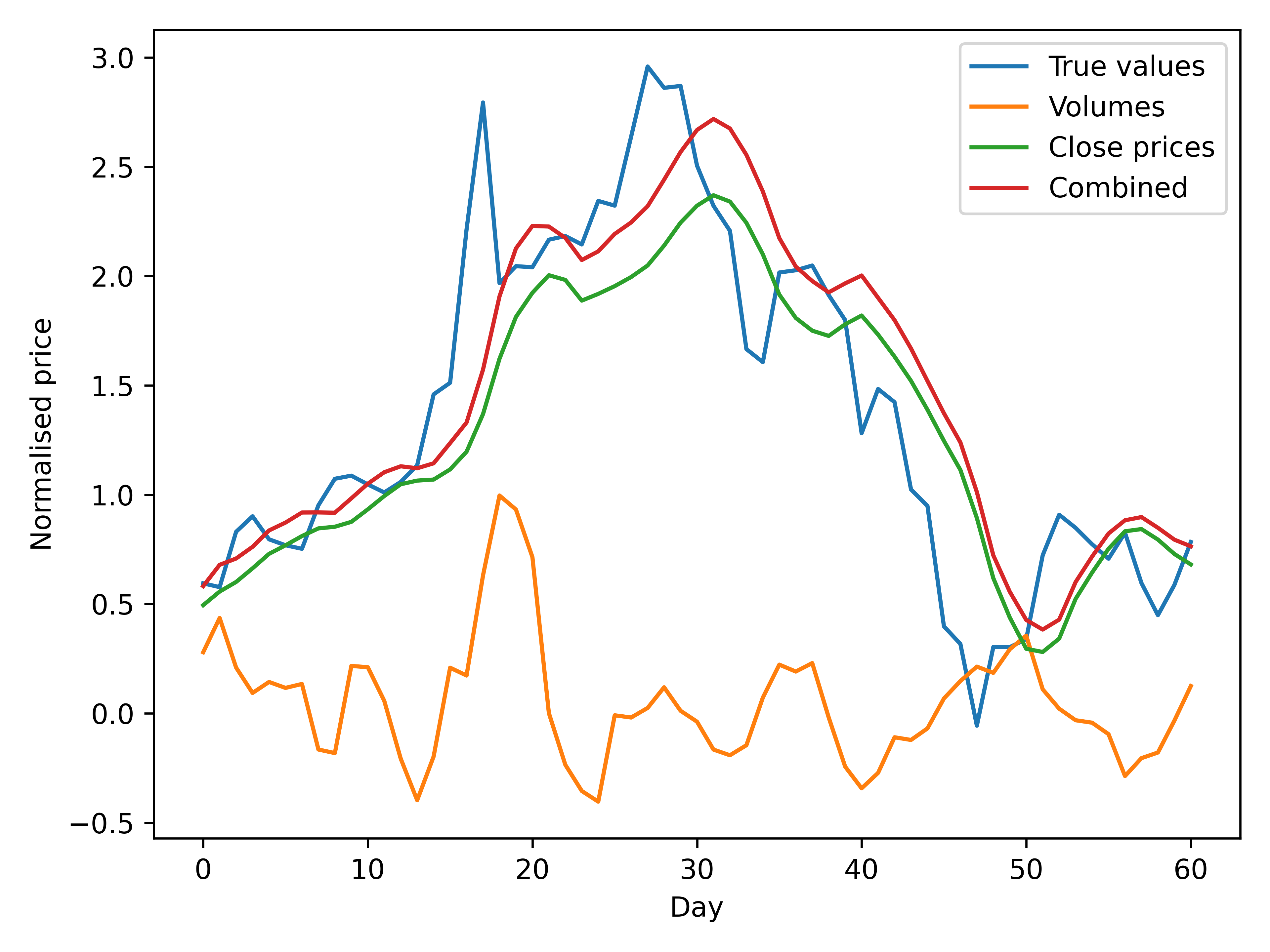}
        \caption{Stock predictions using various features}
        \label{fig:exp2-2-a}
    \end{subfigure}
    ~ 
    \begin{subfigure}[b]{0.4\textwidth}
        \includegraphics[width=\textwidth]{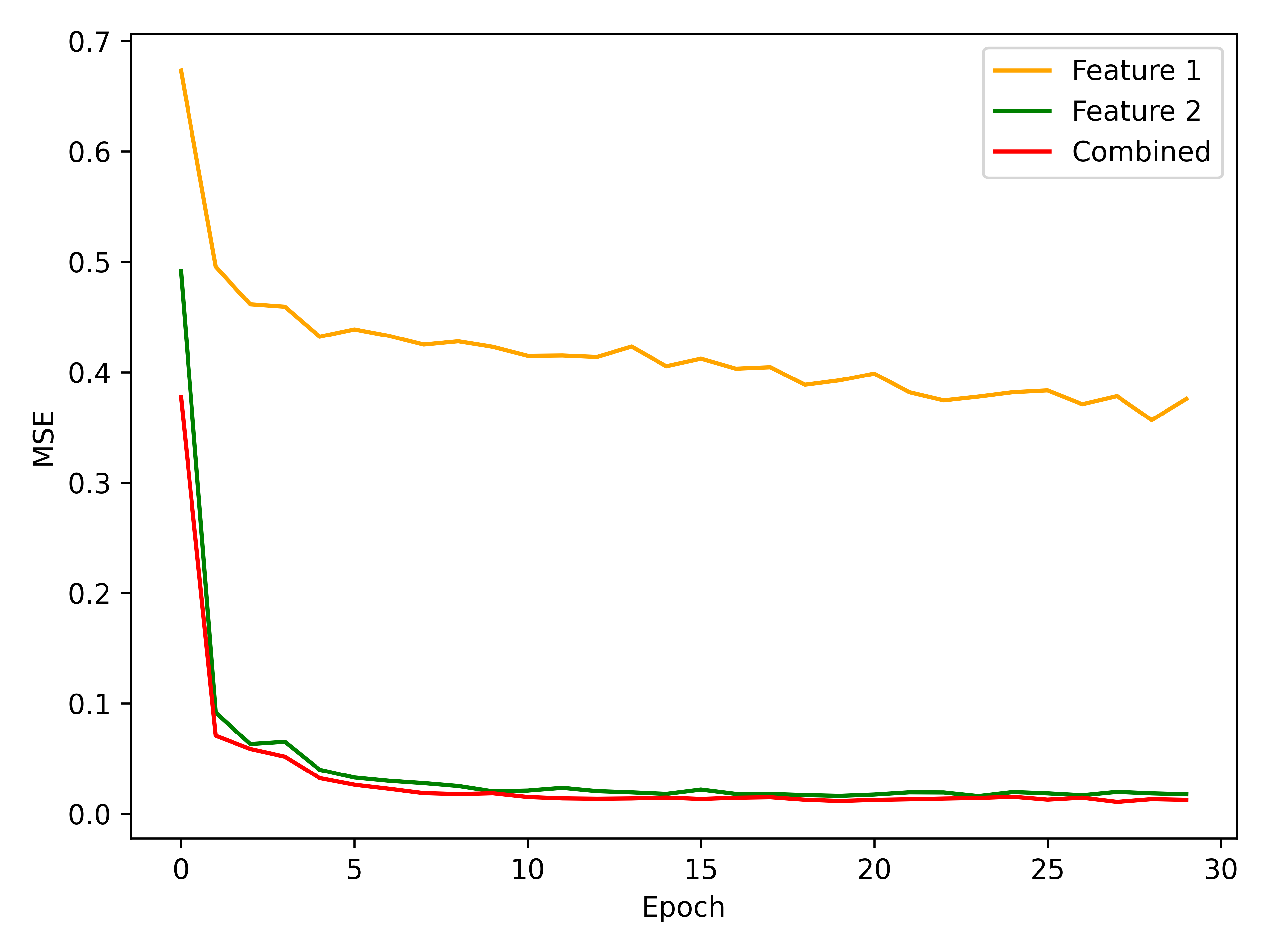}
        \caption{The learning curves of the models}
        \label{fig:exp2-2-b}
    \end{subfigure}
  \end{widepage}
  \caption{The TSLA stock normalised close prices; Loss function and predicted prices using various number of features}
  \label{fig:exp2-2}
\end{figure}

By comparing stock prediction using 1 feature, 2 features and 10 features, we can see that more features generally leads to closer predictions \ref{fig:exp2-2}.

\section{Training of The Neural Network}

\begin{figure}[H]
    \centering
    \begin{subfigure}[b]{0.475\textwidth}
        \centering
        \includegraphics[width=\textwidth]{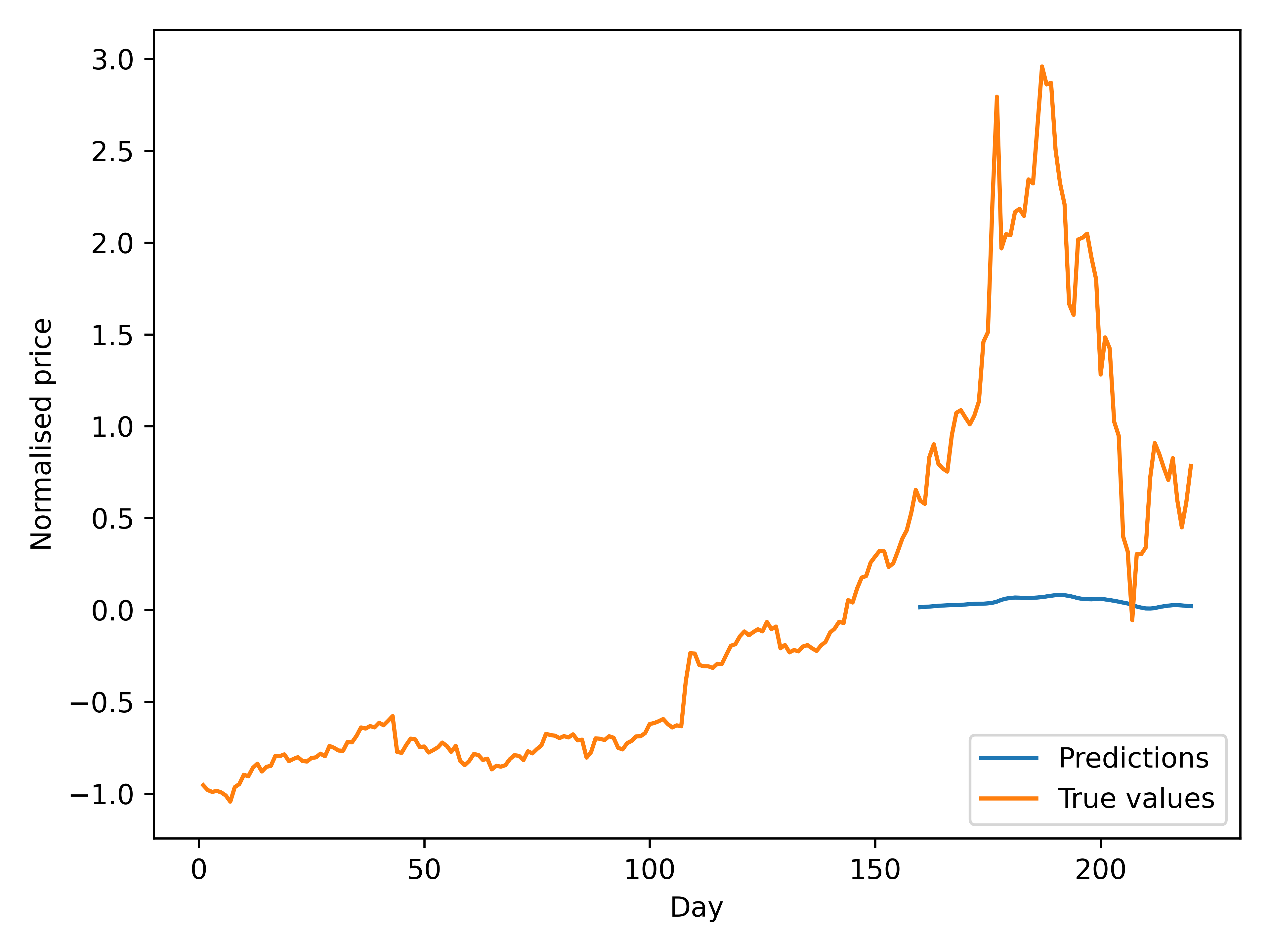}
        \caption{Before training}
    \end{subfigure}
    \hfill
    \begin{subfigure}[b]{0.475\textwidth}
        \centering
        \includegraphics[width=\textwidth]{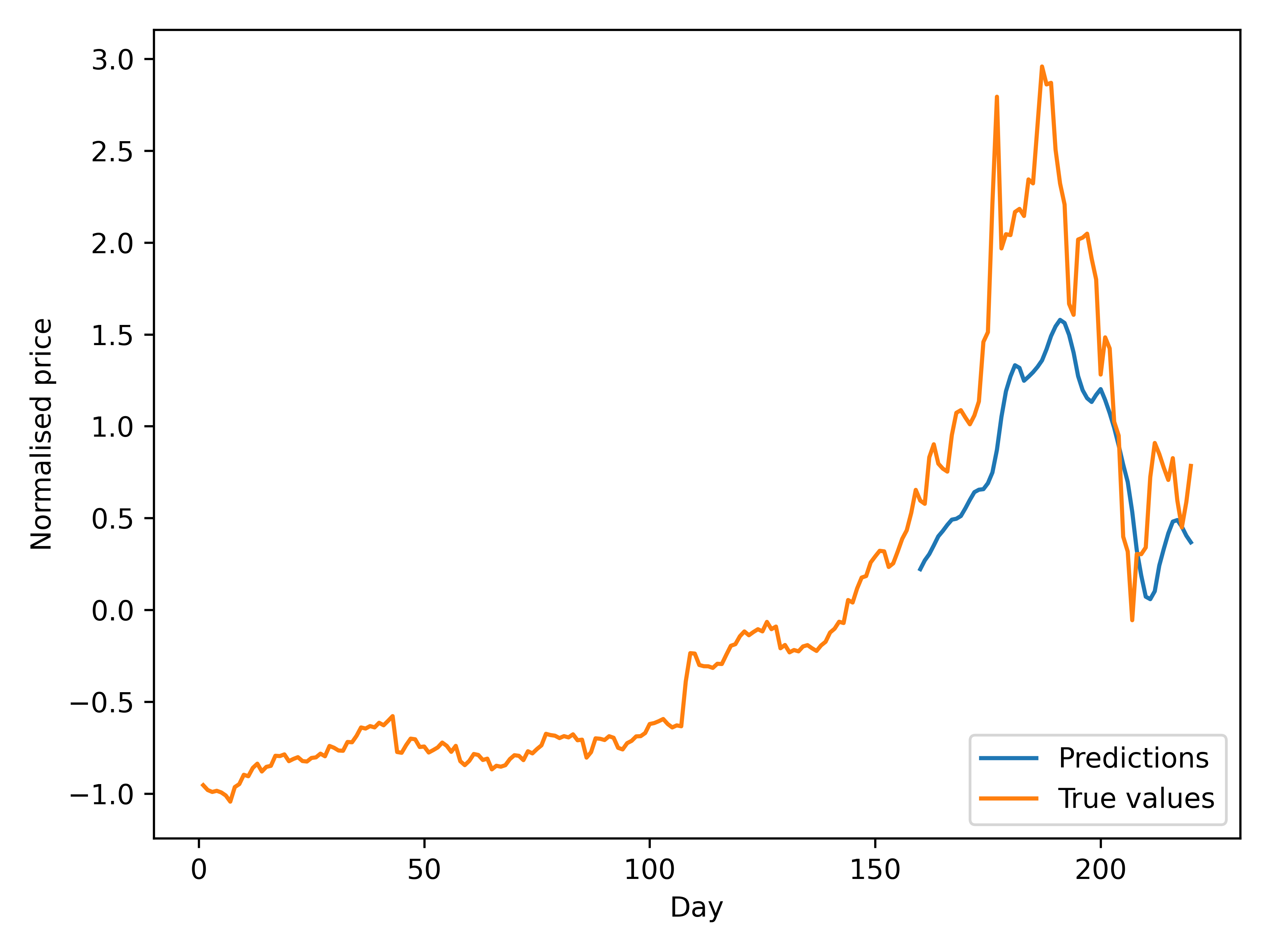}
        \caption{Training for 5 epochs}
    \end{subfigure}
    \vskip\baselineskip
    \begin{subfigure}[b]{0.475\textwidth}
        \centering
        \includegraphics[width=\textwidth]{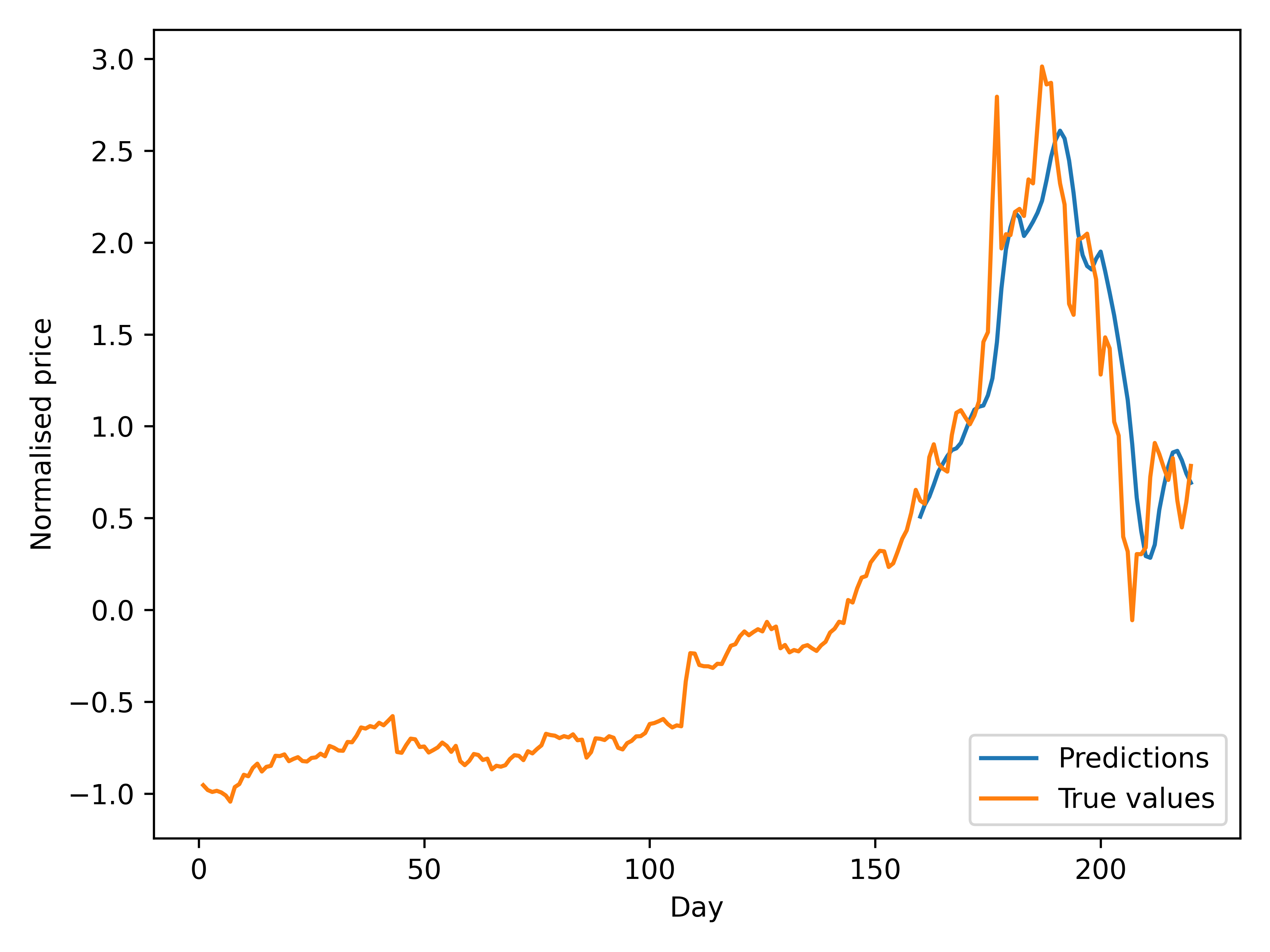}
        \caption{Training for 100 epochs}
    \end{subfigure}
    \hfill
    \begin{subfigure}[b]{0.475\textwidth}
        \centering
        \includegraphics[width=\textwidth]{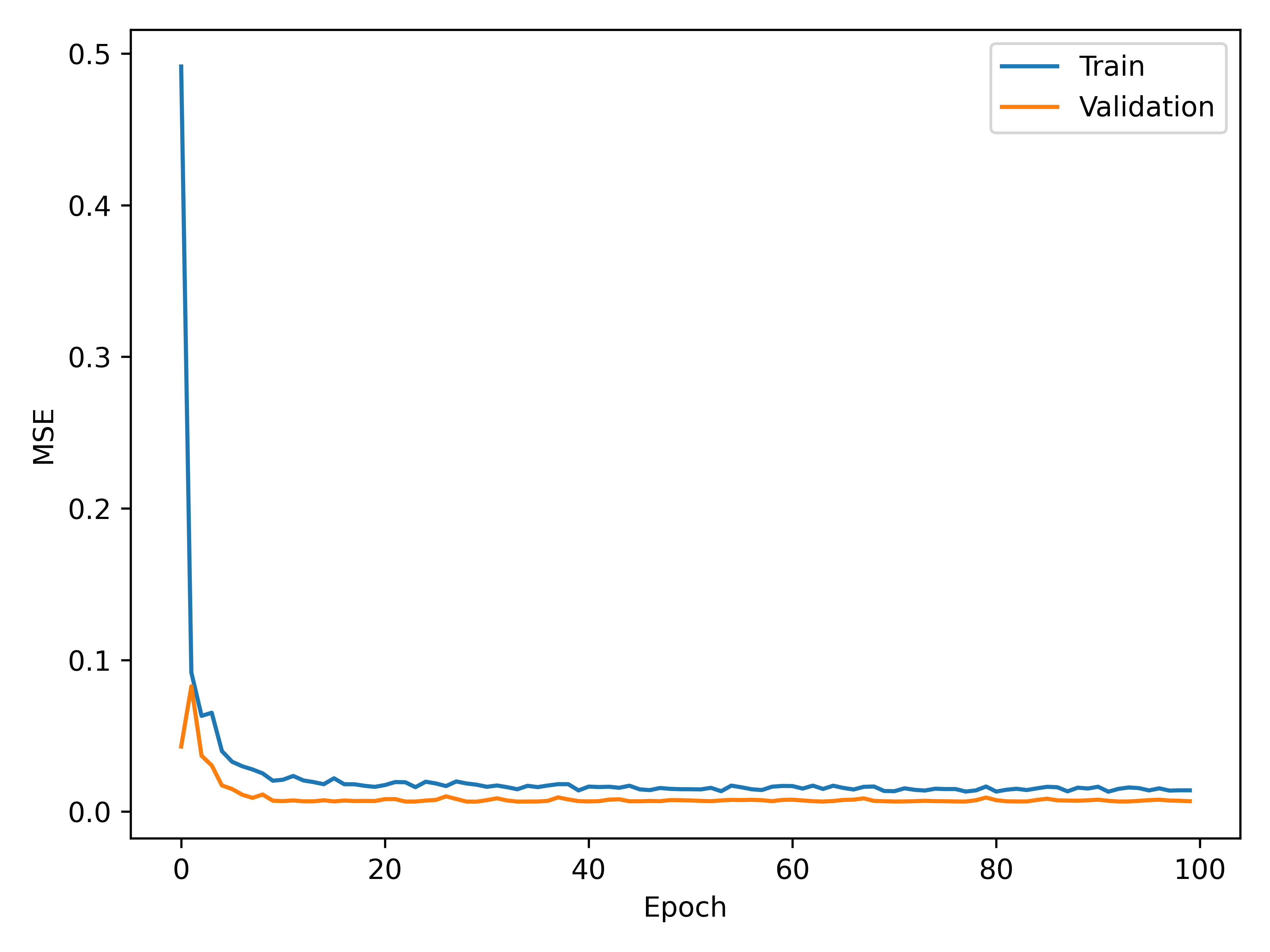}
        \caption{Learning curve}
    \end{subfigure}
  \caption{The predicted stock price during training using the TSLA data set}
  \label{fig:exp1}
\end{figure}

We have designed a neural network consisting of three layers \ref{fig:model} with 200 nodes in the LSTM layers and 200 nodes in the fully connected (dense) layers, with a learning rate of 0.008 and a decay rate of 97\%. The model is trained with the daily close price of the stock TSLA from 23 May 2019 to 6 April 2020. The Tesla data set is split into a training set, a validation set and a test set with percentages 63\%, 7\% and 30\%. Figure \ref{fig:exp1} shows the predicted stock price before training, after training for one epoch (iteration) and after 100 epochs. The first part of the data is used for training and the second part is used for testing. The predictions made in testing are shown as blue in the figure below. The ground truth is shown in yellow just for reference.

\section{Tweet Attributes}

We added the following attributes as features to the data set: sentiment scores, favourites, followers, retweets and verified account. In the first experiment, we used linear scaling to assign weights to them.

\begin{figure}[ht]
  \begin{widepage}
    \centering
    \begin{subfigure}[b]{0.4\textwidth}
        \includegraphics[width=\textwidth]{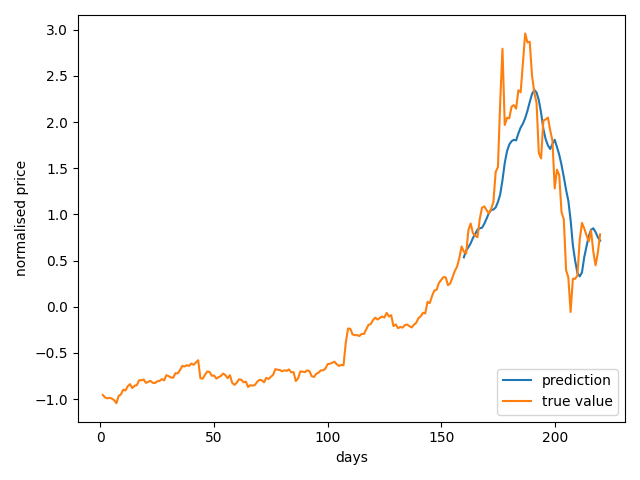}
        \caption{Using only sentiment scores}
    \end{subfigure}
    ~ 
    \begin{subfigure}[b]{0.4\textwidth}
        \includegraphics[width=\textwidth]{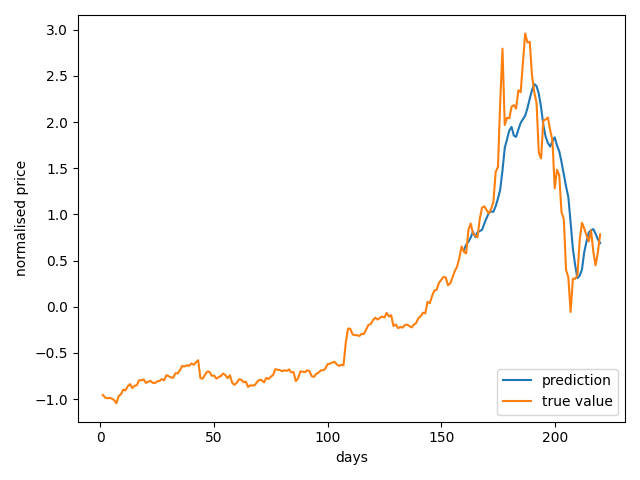}
        \caption{Using sentiment scores and other attributes}
    \end{subfigure}
  \end{widepage}
  \caption{Comparison between additional Twitter attributes}
  \label{fig:exp3-1}
\end{figure}

\begin{figure}[ht]
  \begin{widepage}
    \centering
    \begin{subfigure}[b]{0.4\textwidth}
        \includegraphics[width=\textwidth]{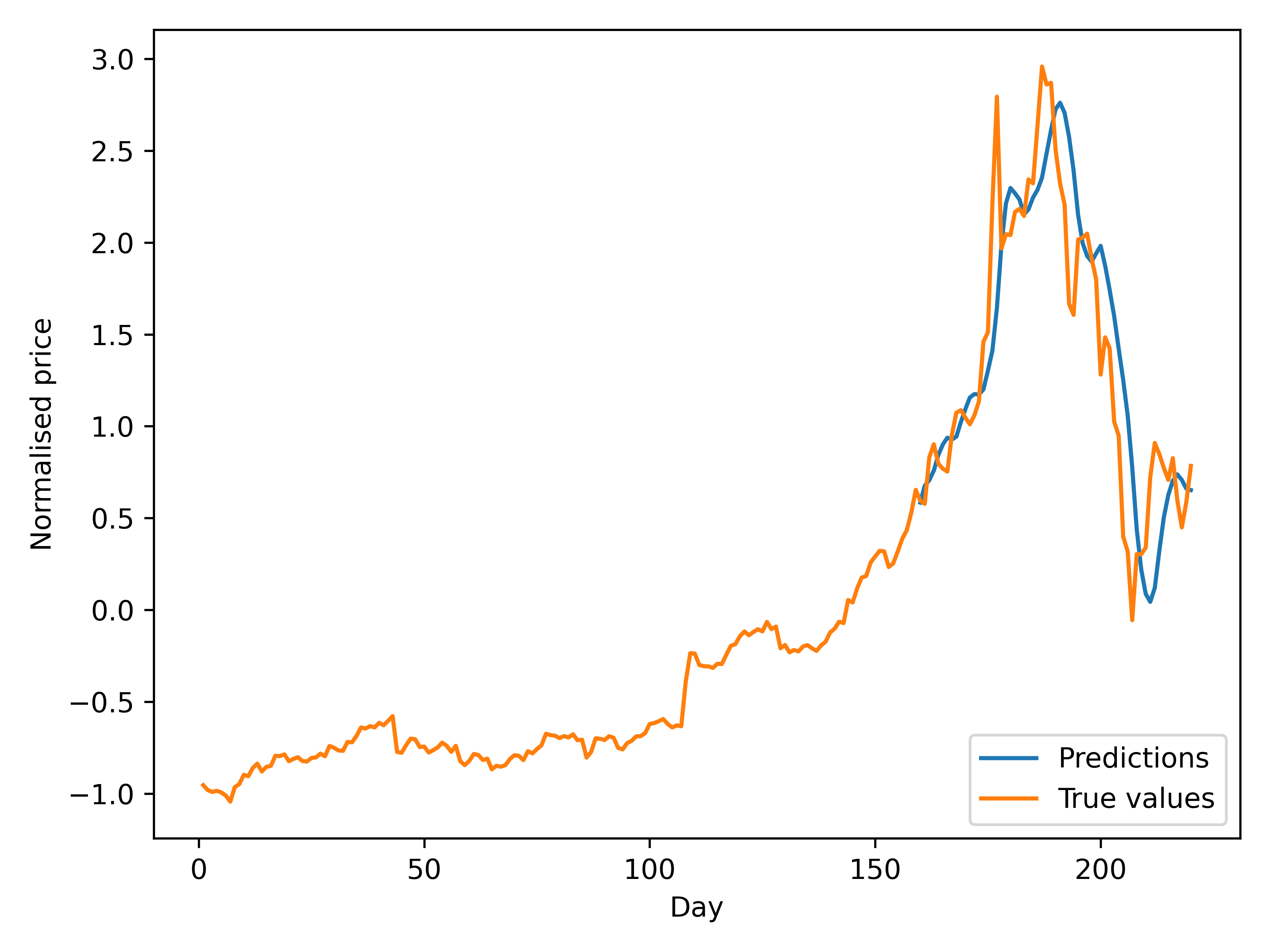}
        \caption{Using only sentiment scores}
    \end{subfigure}
    ~ 
    \begin{subfigure}[b]{0.4\textwidth}
        \includegraphics[width=\textwidth]{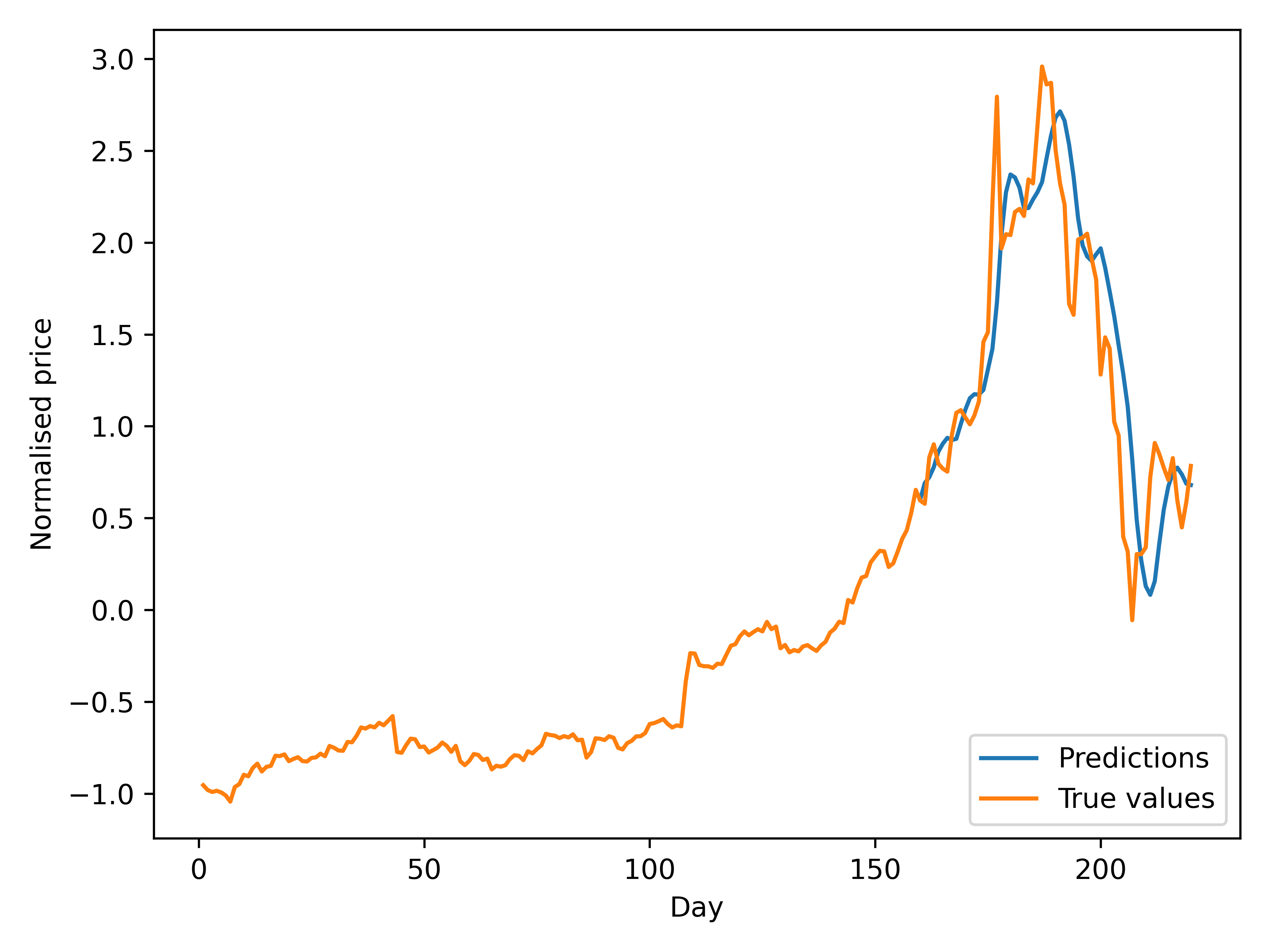}
        \caption{Using sentiment scores and other attributes}
    \end{subfigure}
  \end{widepage}
  \caption{Comparison between stock price prediction using stock data, technical analysis and tweet sentiment scores, and using stock data, technical analysis, tweet sentiment scores and tweet attributes with linear scaling}
  \label{fig:exp3-2}
\end{figure}

We can see that extra Twitter attributes helped to predict more accurate than only sentiment scores \ref{fig:exp3-1}. By experimenting with different weights for the attributes, the best results are presented in figure \ref{fig:exp3-2}.

However, when the network is trained to near its maximum potential, little difference is observed between the model trained with Twitter attributes and without. We do notice 0.002 in MSE and about a 3\% difference improvement.

\begin{figure}[ht]
  \begin{widepage}
    \centering
    \begin{subfigure}[b]{0.4\textwidth}
        \includegraphics[width=\textwidth]{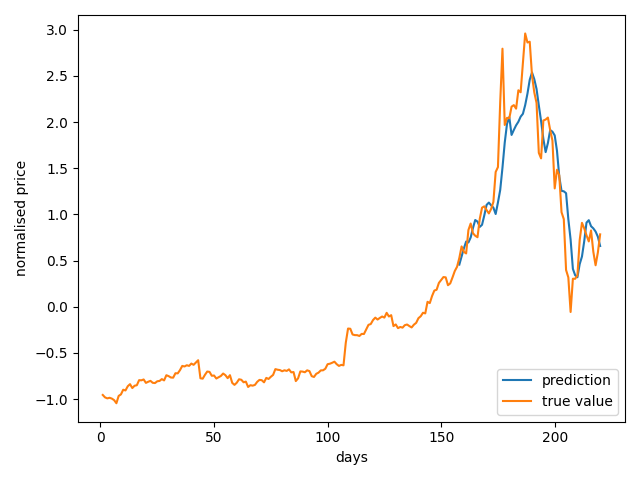}
        \caption{Predicting stock prices using 10 features}
    \end{subfigure}
    \qquad
    ~ 
    \begin{subfigure}[b]{0.4\textwidth}
        \includegraphics[width=\textwidth]{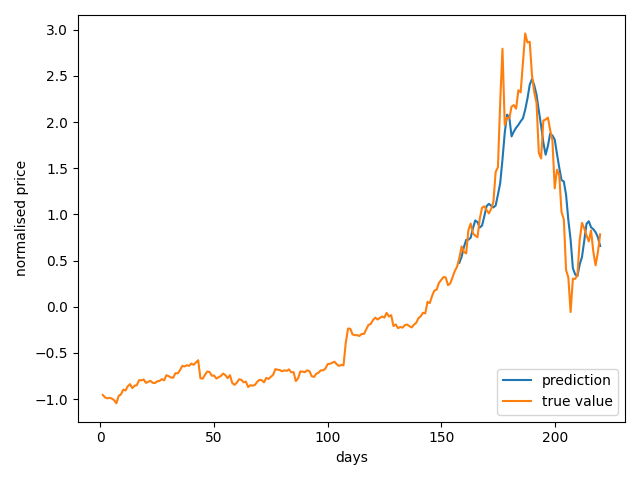}
        \caption{Predicting stock prices using 10 features + tweet attributes with sigmoid scaling}
    \end{subfigure}
  \end{widepage}
  \caption{Comparison between stock price prediction using stock data and technical analysis and using stock data, technical analysis, tweet sentiment scores and tweet attributes}
  \label{fig:exp4-1}
\end{figure}

In the second experiment, we used the sigmoid function on the follower count, as \cite{HongKeelSul_2014} suggested. The results showed that the linear scaling for all attributes gave an MSE of 0.14 on the test set. We also did another experiment with the sigmoid function when scaling the tweet follower count attribute. An MSE of 0.13 was acquired and the comparison is shown in figure \ref{fig:exp4-1}.

\chapter{Discussion}

The research questions for this thesis are:

\emph{Will technical analysis and Twitter attribute information help to predict the stock price?}

If the answer to the question above is yes;

\emph{How much better can a machine learning model become with those mentioned above taken into account?}

By using Tweet attributes along with historical stock prices, sentiment analysis and technical analysis, the MSE (mean squared error) between the predicted prices and the actual prices was successfully brought from 0.1617 down to 0.1437. That is an 11\% improvement. However, when we experimented further to isolate the "tweet attributes" variable, no significant improvement was found. By taken the Tweet attributes into account, the difference was only 3\%.

We have successfully developed a program and collected an enormous amount of tweets, although some were missing due to the limitations of our computation resources and the Twitter API. During the data collection, the API server sometimes returns an error code that crashed the program. This resulted that us do not have any tweets collected for two days. We were also banned from Twitter once during the collection, because of the intense API queries made. We were banned again from Twitter a few days later and the collection came to an end.

The TSLA stock surged from autumn 2019 to an all-time high in early 2020. Then it dropped abruptly to the level before the rally. The training of the neural network model involves splitting the data set into a training set, validation set and test set. Conventionally, the data set is shuffled between the split and the tests are done multiple times to remove potential noises. However, in time-series prediction, the elements in the data set are data points whose order is important. It is possible to shuffle small series of values in the test set, but it would add complexity to the data structures used in this study, to be able to visualise them.

By only shuffling the training set and validation set, the randomness of the inputs was present but the test set is biased. The patterns in the test set are seldom seen in the training set. So, the network may fail to recognise them to predict future prices accurately.

Next, the preprocessing technique used in this thesis may not be optimal for tweet scores. The preprocessing model used is from a well-known python library for data science, \textit{scikit-learn}. The stock data is all in USD but the Twitter sentiment score, for example, can be negative values. This may result in wrong scaling after the preprocessing transformation to the normalised scale. More specifically, if the model fits 300 USD to 3.0 and 100 USD to -1.0, then a sentiment score of 3 would be scaled to a very small number. This could have minimalised the effect of tweet attributes so that they are contributing to better predictions but we do not observe it because of this reason.

Thirdly, the stock market is closed during weekends and US holidays but Twitter is not. During trading days, the tweets are assumed to influence the stock value for the next day. But tweets posted on Friday is influencing the next Monday. As \cite{HongKeelSul_2014} pointed out, Twitter users with fewer followers have an impact on stock prices in the long run and users with more followers affect stock prices mainly intraday. The fact that we used three days of tweets to predict the next trading day's stock price conflicts with the results of their study.

Last but not the least, we have collected tweets for Intel, AMD, Tesla and Microsoft. However, the quality of the other three stocks was not as good as Tesla. It also requires manual work to merge duplicate tweets, as well as tweets during weekends and holidays. Due to the time limit, this part could not be done. We have also planned on adding ARIMA (Autoregressive Integrated Moving Average) to this model, but it was way more complex than we expected.

\section{Conclusion}

We have reached all subgoals and successfully reduced the prediction error by 11\%. The goal of this thesis was achieved. Results showed that technical indicators such as SMA5 and Bollinger Bands did the major contribution. We could also observe that Twitter attributes helped to predict stock prices better by around 3\% in a period, in which the stock fluctuated and was highly unpredictable.

For small investors, 3\% may not be a big deal. But for hedge funds and professional investment banks, this could be a huge profit. They have also more resources to fund research projects. Currently, only roughly 150 examples are used to train the network. If a premium twitter developer subscription is purchased, it would be possible to use all-time historical prices together with tweet attributes. This means that the training examples can be 10 times as many, which would make a difference.

We can therefore conclude that both technical analysis and Twitter attributes will help to predict stock price. The improvements are 11\% and 3\% in MSE with the Tesla data set, given the data collected under the period from 23 May 2019 to 6 April 2020.

\section{Future Work}

Future work may include acquiring all historical tweet by purchasing a premium twitter developer subscription. It is also interesting to add ARIMA to handle the linear part of the data set and let the neural network handle the non-linear part.

\printbibliography

\end{document}